\pdfoutput=1
\documentclass[11pt]{article}
\usepackage[final]{acl}

\usepackage{times}
\usepackage{latexsym}

\usepackage[T1]{fontenc}
\usepackage[utf8]{inputenc}
\usepackage{float}
\usepackage{CJKutf8}
\usepackage{microtype}
\usepackage{amssymb}
\usepackage{svg}
\usepackage{lineno}
\usepackage{tikz}
\usetikzlibrary{arrows.meta}
\usepackage{array}
\usepackage{times}
\usepackage{helvet}
\usepackage{courier}
\usepackage{multirow}
\usepackage{mathrsfs}
\usepackage{graphicx}
\usepackage{enumitem}
\usepackage{graphicx}
\usepackage{blindtext}
\usepackage{algorithm}
\usepackage{algorithmic}
\usepackage[marginal]{footmisc}
\usepackage[utf8]{inputenc}
\usepackage[english]{babel}
\usepackage{epstopdf}
\usepackage{array}
\usepackage{svg}
\usepackage{multirow}
\usepackage{booktabs}
\usepackage{amsmath}
\usepackage{wrapfig}
\usepackage{inconsolata}
\usepackage{url}
\definecolor{lightred}{RGB}{255, 235, 235}
\definecolor{lightgreen}{RGB}{235, 255, 235}

\usepackage{url}
\usepackage{multirow}
\usepackage{svg}

\usepackage{xcolor}
\usepackage[table]{xcolor} 
\definecolor{lightpink}{HTML}{ed9782}
\definecolor{lightblue}{HTML}{5395f5}
\definecolor{lightgreen}{HTML}{efd08f}
\definecolor{grey}{HTML}{b3b3b3}
\definecolor{cgreen}{RGB}{46, 139, 87} 

\definecolor{lightorange}{RGB}{255, 190, 123}
\usepackage[most]{tcolorbox}

\usepackage{algorithm}
\usepackage{algorithmic}

\usepackage{enumitem}
\usepackage{tabularx}
\usepackage{subcaption}
\usepackage{makecell}

\usepackage{tabularx}
\usepackage{multirow}
\usepackage[normalem]{ulem}
\useunder{\uline}{\ul}{}

\usepackage{times}
\usepackage{moreverb}
\usepackage{graphicx}
\usepackage{mathrsfs}
\usepackage{bm}
\usepackage{url}
\usepackage{amsmath}
\usepackage{graphicx}
\usepackage{fontawesome5}

\title{\faCrosshairs~Freeze Deep, Train Shallow: Interpretable Layer Allocation for Continued Pre-Training}

\author{
  Yu-Hang Wu\textsuperscript{1,2},
  Qin-Yuan Liu\textsuperscript{1},
  Qiu-Yang Zhao\textsuperscript{1},
  Bo Jiang\textsuperscript{1},
  Jiang-Feng Yang\textsuperscript{1},
  Qing-Wei Chong\textsuperscript{1}\thanks{Corresponding author.}
  \\
  \textsuperscript{1} Nanhu Research Institute of China Electronic Science and Technology
  \\
  \textsuperscript{2} School of Electronic and Electrical Engineering, Shanghai University of Engineering Science
}

\usepackage{wrapfig}  
\usepackage{graphicx} 
\usepackage{subcaption}
\usepackage{balance}
\begin{document}

\maketitle
\begin{abstract}
Selective layer-wise updates are essential for low-cost continued pre-training of Large Language Models (LLMs), yet determining which layers to freeze or train remains an empirical black-box problem due to the lack of interpretable guidance. To address this issue, we propose LayerTracer, an architecture-agnostic diagnostic framework that reveals the evolution patterns of layer-wise representations and stability by locating task execution positions and quantifying layer sensitivity. Analysis results reveal that deep layers act as critical regions for task execution and maintain high stability against disruptive updates. Guided by this finding, we conduct three controlled continued pre-training trials to compare diverse freeze-train strategies, demonstrating that training shallow layers while freezing deep layers consistently outperforms full-parameter fine-tuning and the opposite allocation on both C-Eval and CMMLU benchmarks. We further present a hybrid model case study, which validates that placing high-quality pre-trained modules in deep layers effectively preserves inherent knowledge of the model. This work delivers a low-cost and interpretable solution for resource-constrained teams, offering actionable guidance for layer-wise parameter allocation in continued pre-training and hybrid model construction.
\end{abstract}

\begin{figure}[!t]
    \centering
    \includegraphics[width=\columnwidth]{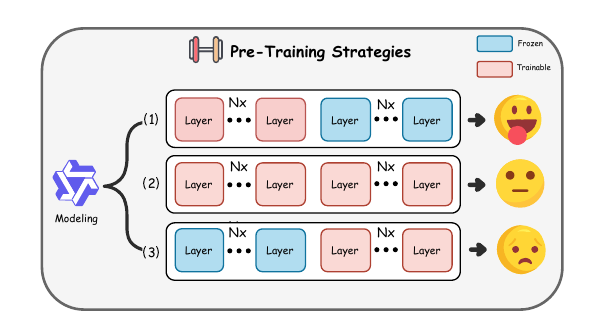}
    \caption{Comparison of three layer-wise pre-training strategies on the CEval dataset~\cite{huang2023ceval} using the Qwen3-0.6B architecture. (1) (Freeze-Deep/Train-Shallow) achieves the best performance, followed by (2) (Full Training), while (3) (Freeze-Shallow/Train-Deep) yields the worst results.}
    \label{fig:example}
\end{figure}

\section{Introduction}
With the rapid development of Large Language Models (LLMs), leading general-purpose models such as GPT~\cite{openai:23gpt4}, LLaMA~\cite{touvron:23llama2,touvron:23llama,dubey:24llama3}, and Qwen~\cite{qwen25:24,qwen2:24,yang:25qwen3} have achieved significant breakthroughs in semantic understanding, logical reasoning, and generation, thereby accelerating industrial adoption and engineering deployment~\cite{zhao:23survey,liu:23robustness,abo2025survey,wu2025sugar}. However, the superior performance of these models relies heavily on massive high-quality pretraining corpora and large-scale computing clusters, creating substantial technical barriers~\cite{bai2023qwen,touvron:23llama2}. For small and medium-sized teams with limited computational power and data reserves, pretraining general-purpose models from scratch is largely infeasible. Consequently, a mainstream paradigm for vertical domain modeling has emerged. This approach leverages the stacked decoder layers of open-source pretrained models by freezing a portion of the decoder layers, while only training the remaining native layers as well as newly added custom layers. Combined with high-quality domain-low-budget continued pre-training data for domain adaptation, this strategy enables the rapid deployment of commercially viable vertical domain models at minimal computational and data costs~\cite{bao2023disc,chen2023meditron,roziere2023code,chen2023huatuogpt,labrak2024biomistral}.

Although this approach has established a standardized pipeline, it suffers from a critical limitation: the lack of interpretable guidance for layer-wise freeze–train allocation during continued pre-training. Thus, small and medium-sized teams are forced to rely on empirical heuristics when determining which layers to freeze or train, leaving the allocation process largely opaque. This reliance on trial-and-error often leads to undesirable overwriting of pre-trained knowledge and performance fluctuations, while substantially increasing computational costs. As illustrated in Figure~\ref{fig:example}, the specific arrangement of frozen and trainable layers critically dictates convergence stability and final performance. Therefore, uncovering functional differentiation patterns across layers and deriving actionable freeze-train rules emerges as a critical imperative. Specifically, we aim to address the following two research questions:

\begin{itemize}
    \item \textit{RQ1: What are the functional differentiation patterns across layers in pretrained models?}
    \item \textit{RQ2: How can these patterns inform practical layer-wise freeze/train allocation strategies for continued pre-training?}
\end{itemize}

\begin{figure}[!t]
    \centering
    \includegraphics[width=\columnwidth]{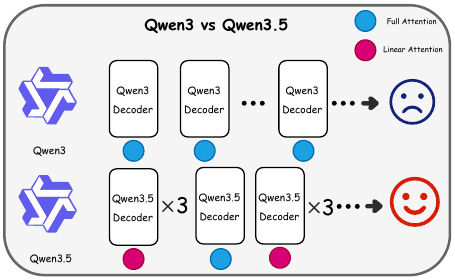}
    \caption{The architectures of Qwen3 model and Qwen3.5 model. Qwen3 adopts a single architecture, while Qwen3.5 is a hybrid architecture with a 3:1 ratio of Linear Attention to Full Attention.}
    \label{fig:background}
\end{figure}

Nevertheless, existing research offers limited guidance for layer-wise parameter allocation in continued pre-training. Although prior layer probing studies~\cite{belinkov2022probing,gurnee2023finding,ju2024largelanguagemodelsencode,Xiao2025analyzing,eisenstadt2025overclockingllmreasoningmonitoring,zhang2026locatesteerimprovepractical} and parameter-efficient tuning methods~\cite{bao2023disc,liu:23robustness} have yielded valuable insights into internal model dynamics and fine-tuning strategies, they mainly focus on interpreting model behaviors or task-specific fine-tuning, rather than producing actionable rules for layer freeze-train decisions. More importantly, few existing efforts develop quantitative paradigms to characterize intrinsic layer-wise properties, resulting in a lack of interpretable principles to address the core research questions \textit{RQ1} and \textit{RQ2}.

To address these issues, we propose LayerTracer, a diagnostic framework for layer-wise freeze/train allocation in continued pre-training. It quantifies two core metrics: \textit{Task Particle (TP)} identifies the layers where task probability undergoes meaningful relative shifts, marking where task evidence actively consolidates. \textit{Layer-wise Sensitivity (LS)} measures the relative change in Jensen-Shannon divergence~\cite{lin:91entropy} across consecutive layers under controlled perturbation, capturing zones Sensitive to disruptive updates. Across the Qwen3 series, we observe a consistent finding: shallow layers exhibit higher sensitivity to perturbations, while deep layers consolidate task evidence and stabilize execution. Based on this finding, we derive a practical allocation rule: freeze deep pre-trained layers and train shallow ones. We validate this rule through three controlled continued pre-training experiments, complemented by a hybrid architecture case study to simulate its practical value in resource-constrained industrial scenarios. Results show that Our train-shallow/freeze-deep strategy achieves a notable relative improvement over the reverse allocation, with an average gain of 15.72\% across C-Eval~\cite{huang2023ceval} and CMMLU~\cite{li-etal-2024-cmmlu}.

\section{Related Work}
\label{sec:related}

\subsection{Evolution of LLM Architectures}
The Transformer~\cite{vaswani:17attention} remains the dominant backbone across modern foundation models~\cite{openai:23gpt4,touvron:23llama,touvron:23llama2,dubey:24llama3,yang:25qwen3}. To overcome the scalability bottleneck of self-attention, efficient alternatives such as State Space Models~\cite{gu:24mamba,dao:24ssm}, Linear Attention~\cite{ahn:24linear}, and GatedDeltaNet~\cite{yang:25gateddelta} have been proposed, giving rise to hybrid systems like Jamba~\cite{lieber:24jamba} and Qwen3.5~\cite{qwen35:25repo} that interleave full-attention and efficient layers as illustrated in Figure~\ref{fig:background}. Such designs enable low-cost domain adaptation by freezing pre-trained foundation layers while tuning lightweight modules~\cite{bao2023disc,chen2023meditron,roziere2023code,chen2023huatuogpt,labrak2024biomistral}. Nevertheless, layer-wise freeze–train decisions remain purely heuristic. Without quantitative and interpretable guidance, improper allocation easily breaks knowledge coherence and degrades robustness, especially in fragile hybrid architectures.

\subsection{Layer-Wise Representation Analysis}
Layer-wise interpretability has evolved from early linguistic probing~\cite{pimentel2020information,belinkov:22probing,youssef2023give} to advanced mechanistic analysis that locates task-critical components and knowledge boundaries~\cite{gurnee2023finding,meng:22locating,geva:21ffn,ju2024largelanguagemodelsencode,Xiao2025analyzing}. Closely related are logit/tuned lens and activation patching and causal tracing techniques~\cite{meng:23mass,hernandez:24linearity,liu:24whatprobes}, which project intermediate states through the LM head or intervene on activations to reveal internal dynamics. However, these works focus on post-hoc interpretation rather than actionable guidance. They lack quantitative metrics for layer robustness and cannot be directly generalized to hybrid models. No prior framework unifies task localization and sensitivity measurement for continued pre-training.

\subsection{Parameter-Efficient Adaptation and Allocation}
Parameter-efficient tuning methods such as LoRA~\cite{hu2021loralowrankadaptationlarge}, AdaLoRA~\cite{zhangadaptive}, and DCFT~\cite{zhang-etal-2025-parameter} optimize layer-wise allocation for fine-tuning, with extensions to instruction tuning and alignment via RLHF~\cite{schulman2017proximal,rafailov2023direct,guo2025deepseek}. Yet these approaches rely on task-specific gradients or labeled data, making them unsuitable for unsupervised continued pre-training on raw text. Existing strategies also ignore intrinsic layer stability and fail to generalize to hybrid architectures. This creates a critical need for a gradient-free, interpretable diagnostic to guide principled layer allocation.

\begin{figure*}[t]
    \centering
    \includegraphics[width=\textwidth]{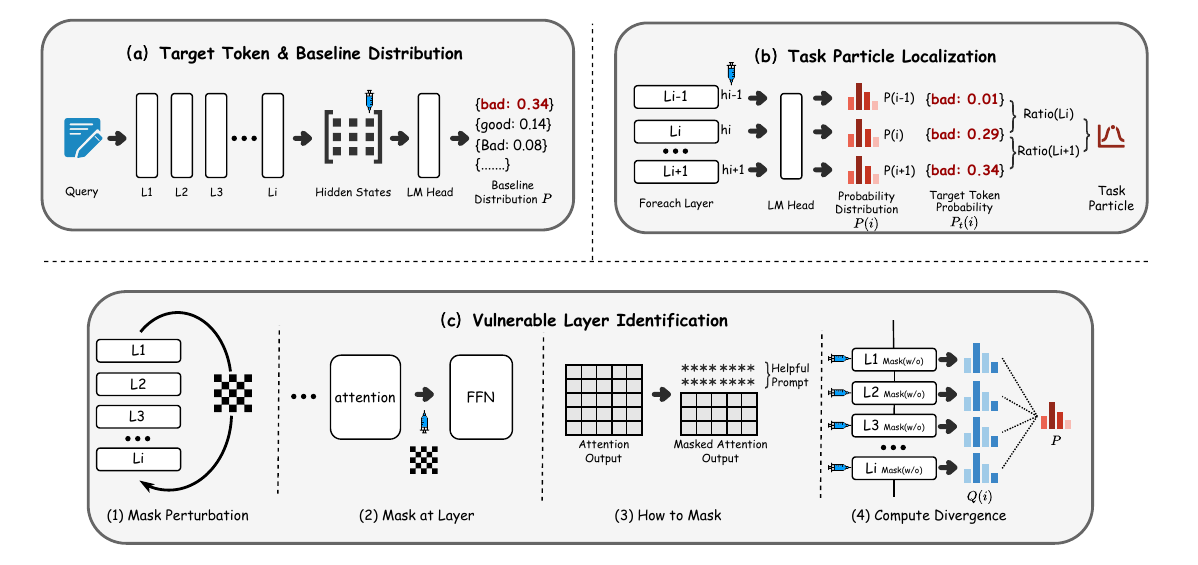}
    \caption{Overview of the LayerTracer framework. 
    (a) Baseline projection: hidden states at each layer are projected via the shared LM head to obtain the target token probability $P_t(l)$, with $t^*$ selected from the final distribution $P$. 
    (b) Task Particle: we compute the relative probability shift $\mathrm{Ratio}(l)$ across consecutive layers. Layers satisfying $\mathrm{Ratio}(l) > 0$ form a execution interval, marking depths where task evidence actively consolidates. 
    (c) Layer-wise Sensitivity: we apply context-targeted masking perturbations at layer $l$ and compute the relative fluctuation $\Delta\mathrm{JS}(l)$ of JS divergence across adjacent layers, identifying zones highly sensitive to parameter updates and information flow disruptions.}
    \label{fig:framework}
\end{figure*}

\section{Method}
\label{sec:method}
To answer \textit{RQ1} and \textit{RQ2} raised in this paper, this section proposes a hierarchical layer analysis framework named LayerTracer. The overview of the LayerTracer framework is illustrated in Figure~\ref{fig:framework}.

\subsection{Preliminary}
\label{sec:preliminary}

We first define unified notations for layer-wise analysis. The input follows a composite task structure \( t = s_1 \boldsymbol{\oplus} s_2 \), where \( s_1 \) denotes context examples and \( s_2 \) denotes the query input. Given such structured input, the model outputs a series of hidden states $\{h_1, h_2, \dots, h_N\}$, where $h_l$ represents the hidden state at the $l$-th layer. All layer-wise hidden states are projected to the vocabulary space via the shared final-layer LM head, ensuring consistent distribution calibration and controlled variables across layers. 

The final output probability distribution $P$ is obtained by projecting the last-layer hidden state $h_N$ through the shared LM head. We define the target token $t^*$ as the token with the maximum probability in $P$. This token is determined by the model's own final prediction rather than external labels, and is used to trace how the model's preferred answer emerges across layers. Let $P_t(l)$ denote the probability of $t^*$ derived from the hidden state $h_l$ projected by the shared LM head. For brevity, we refer to $P_t(l)$ as task probability in subsequent analysis, and the collection of layers with meaningful $P_t(l)$ shifts as task evidence consolidation zones. For a masking perturbation applied at layer $l$, we mask only the context $s_1$ while keeping the query $s_2$ unchanged and use $Q(l)$ to denote the final output distribution after perturbation. The Jensen-Shannon divergence between two distributions $P$ and $Q$ is denoted as $\mathrm{JS}(P \parallel Q)$.

\subsection{Task Particle}
\label{sec:task_particle}
The \textit{Task Particle (TP)} is designed to locate the critical layer where the model initiates task execution. Following the notation in section~\ref{sec:preliminary}, for each layer $l$, we compute the probability of the target token $t^*$ and measure its relative increase ratio between consecutive layers:
\begin{equation}
\mathrm{Ratio}(l) = \left| \frac{P_t(l) - P_t(l-1)}{P_t(l) + \epsilon} \right|,
\label{eq:ratio_definition}
\end{equation}
where $\epsilon = 10^{-6}$ ensures numerical stability, and $P_t(l-1)$ denotes the task probability from the preceding layer. The Task Particle is formally defined as the set of layers where $\mathrm{Ratio}(l) > 0$, forming a continuous task-execution interval. Unlike single-point localization methods, this interval-based view captures all depths where task evidence undergoes meaningful relative shifts, encompassing both task probability promotion and suppression. This continuous zone marks the region where the model actively consolidates and executes task-relevant representations.

\begin{figure*}[t]
    \centering
    \subfloat[Qwen3-0.6B-Base]{\label{fig:a}
        \includegraphics[width=0.32\textwidth]{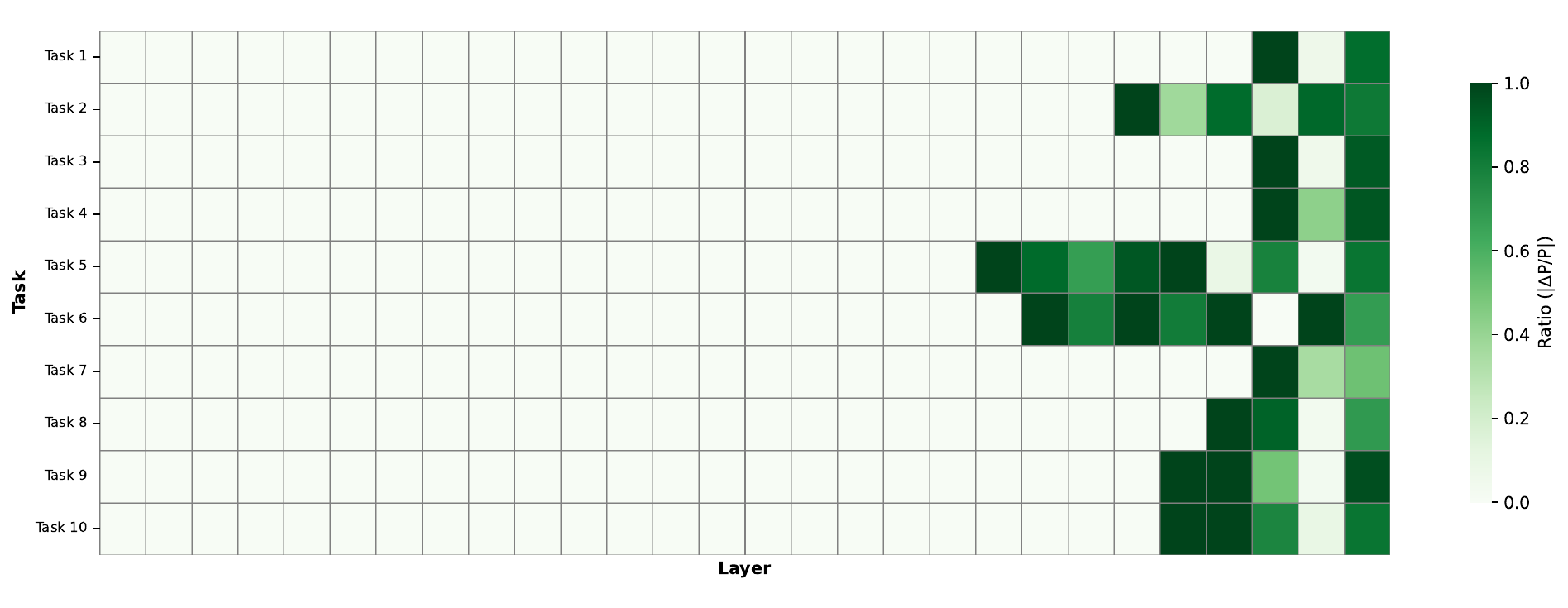}}
    \hfill
    \subfloat[Qwen3-8B-Base]{\label{fig:b}
        \includegraphics[width=0.32\textwidth]{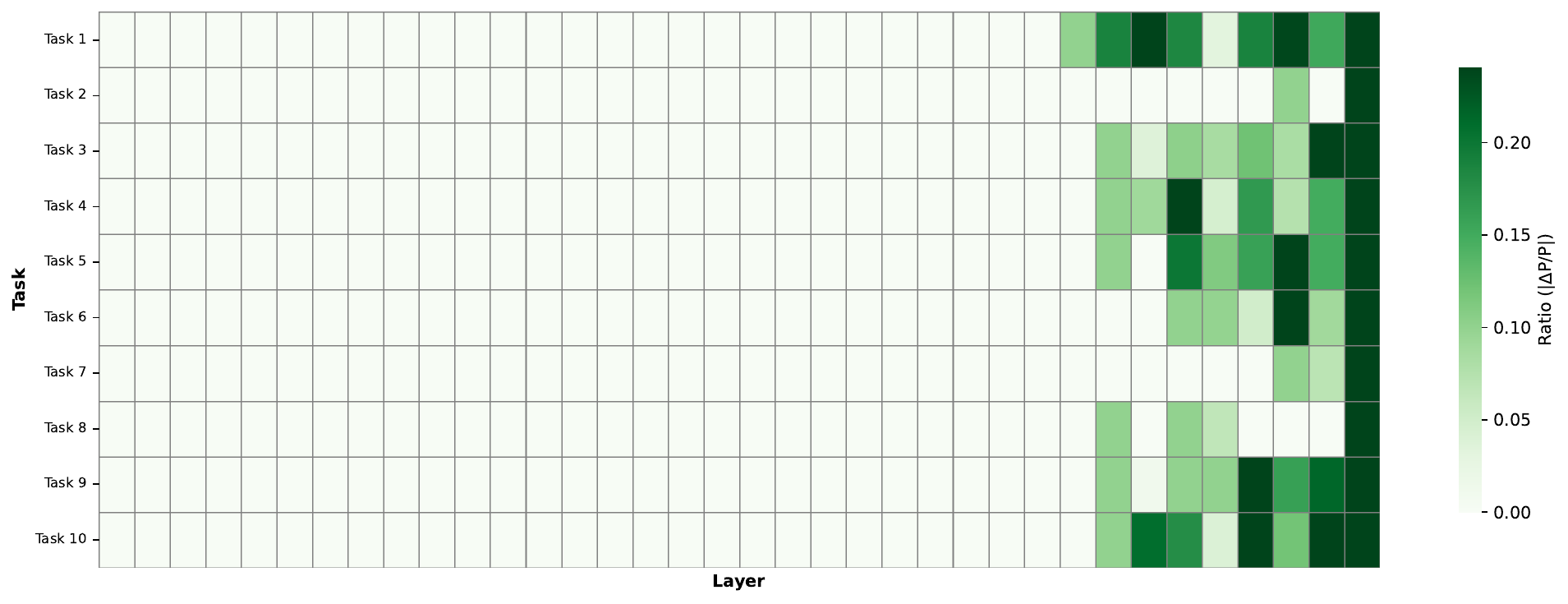}}
    \hfill
    \subfloat[Qwen3-14B-Base]{\label{fig:c}
        \includegraphics[width=0.32\textwidth]{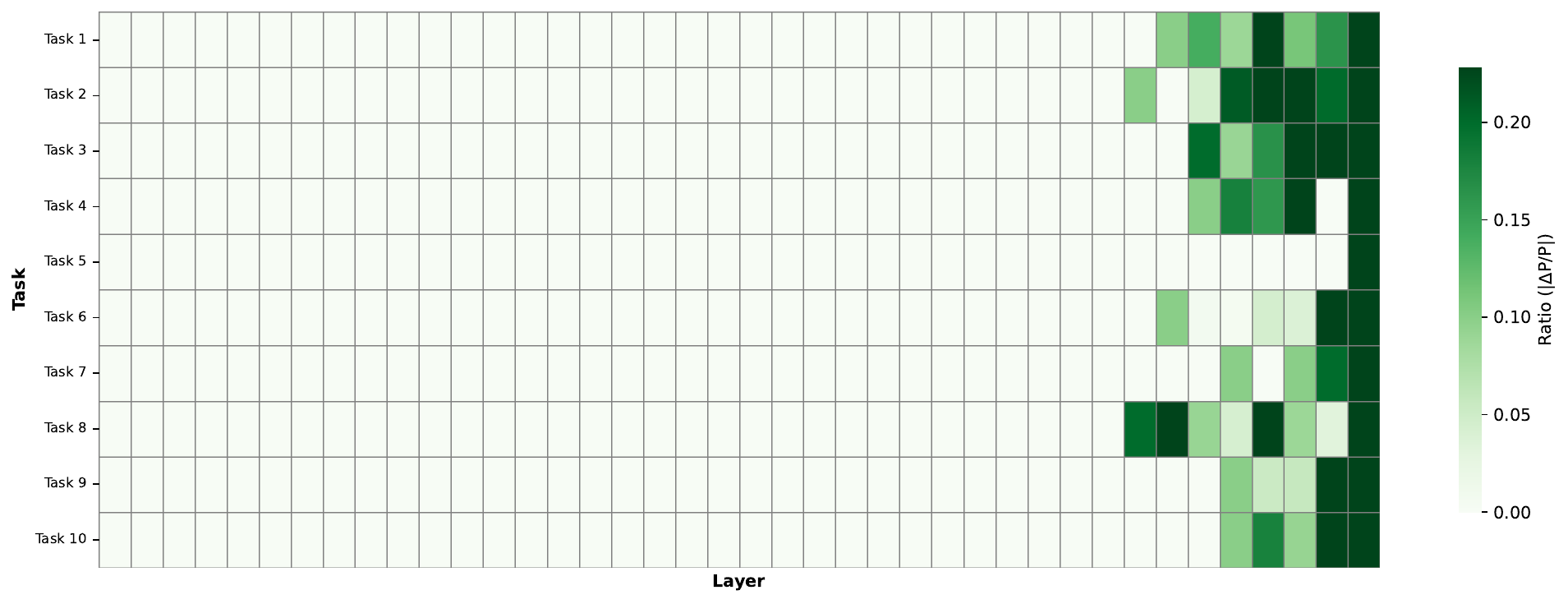}}

    \subfloat[Qwen3-0.6B-Base]{\label{fig:d}
        \includegraphics[width=0.32\textwidth]{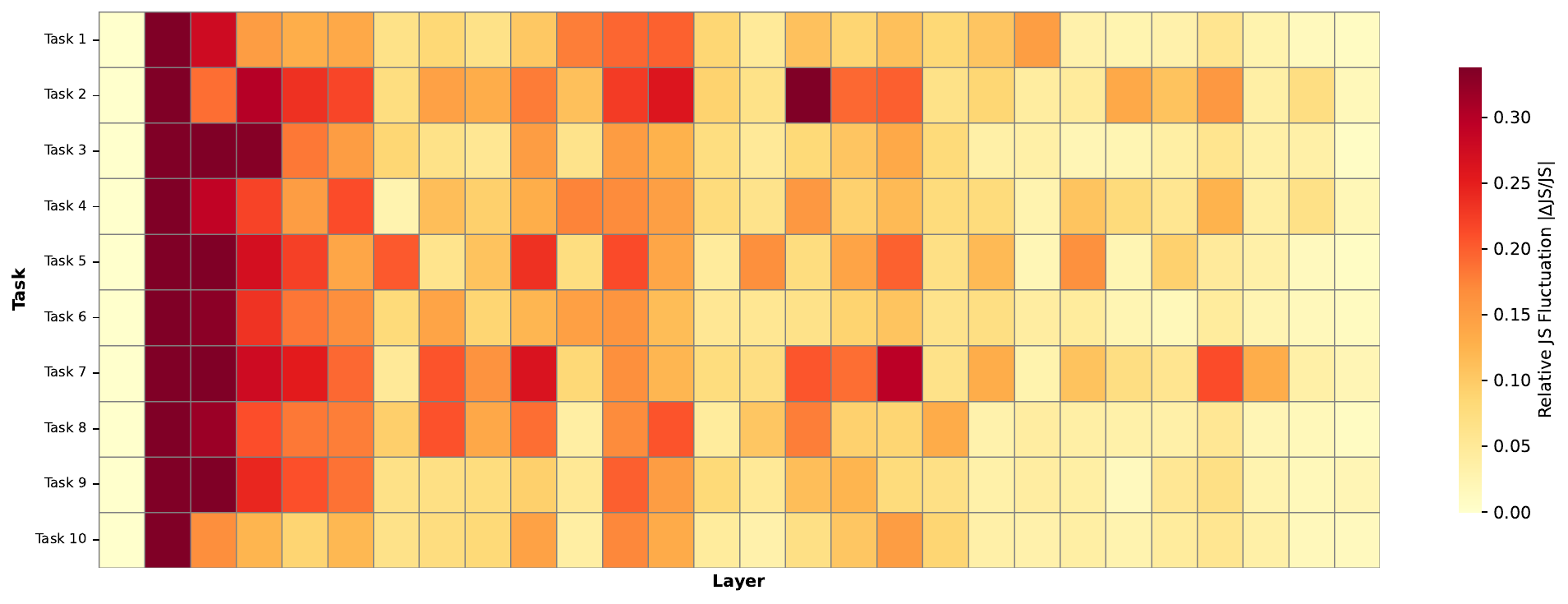}}
    \hfill
    \subfloat[Qwen3-8B-Base]{\label{fig:e}
        \includegraphics[width=0.32\textwidth]{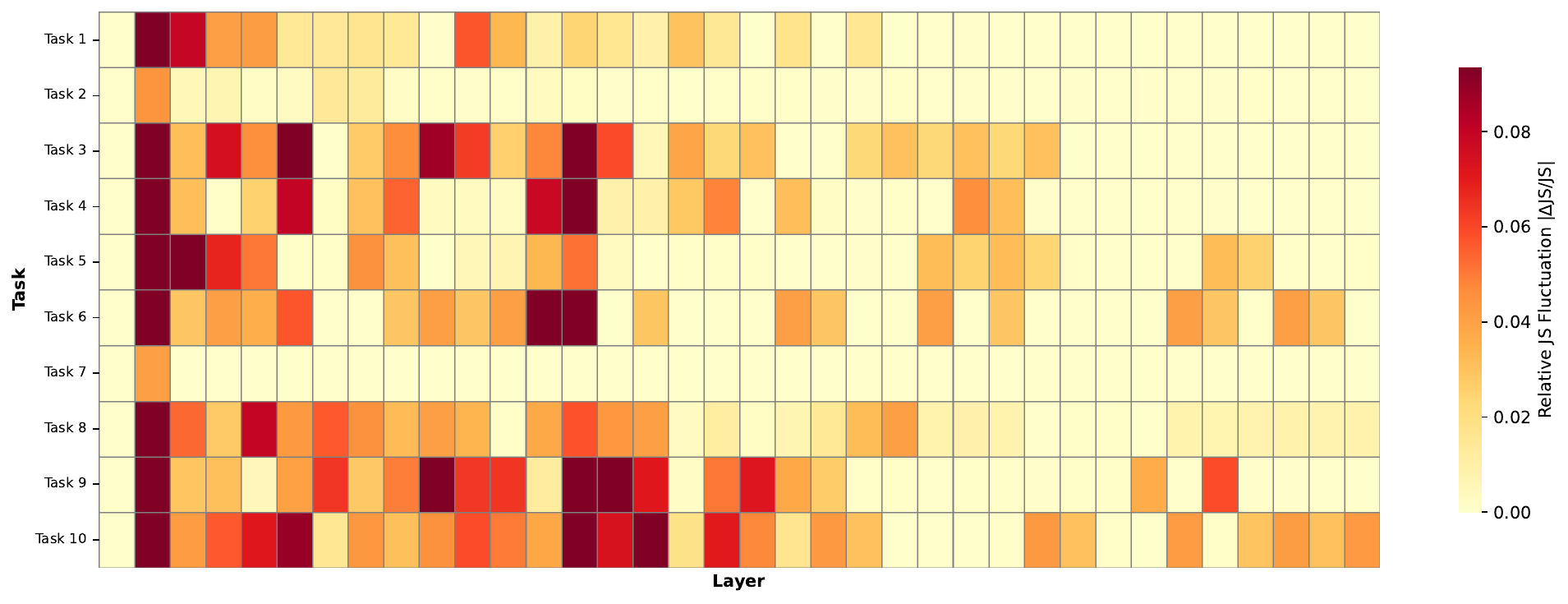}}
    \hfill
    \subfloat[Qwen3-14B-Base]{\label{fig:f}
        \includegraphics[width=0.32\textwidth]{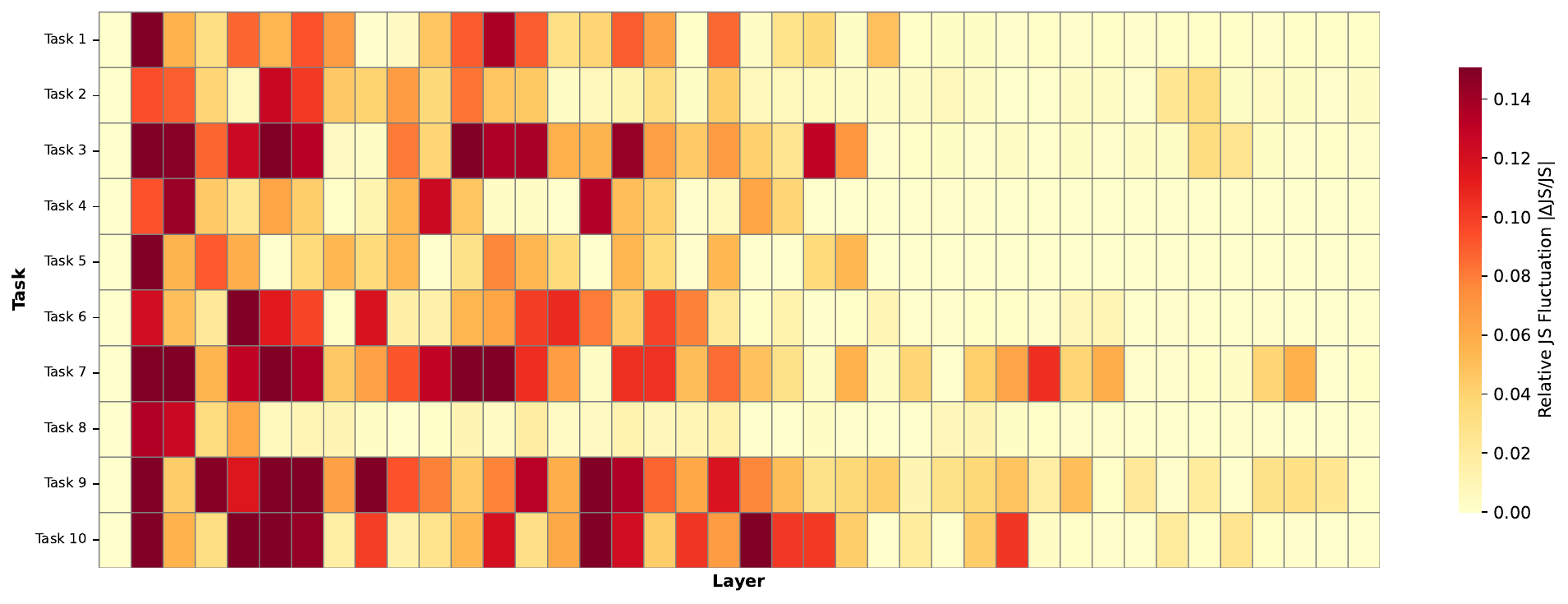}}

    \caption{
        Distribution details in the AntSynNET dataset, where 500 samples are evenly divided into 10 groups of 50 samples, denoted as Task~1--Task~10 for visualization.
        (a)--(c) present the variation of $\mathrm{Ratio}$ across different layers for Qwen3-0.6B-Base to Qwen3-14B-Base.
        A deeper green color indicates a larger $\mathrm{Ratio}$, meaning that the target token probability changes more significantly at this layer, marking a key position for task execution.
        (d)--(f) present the variation of $\Delta\mathrm{JS}$ across different layers, which measures the relative fluctuation of JS divergence between adjacent layers.
        A deeper color indicates a larger abrupt change in layer sensitivity, reflecting higher vulnerability and lower robustness.
        }

    \label{fig:main_results} 
\end{figure*}

\subsection{Layer-wise Sensitivity}
\label{sec:sensitivity}

To evaluate layer-wise stability under perturbations, we propose the Layer-wise Sensitivity (LS) metric.
Different from layer-level evaluation that directly adopts raw JS divergence, LS focuses on the relative variation trend across consecutive layers, thereby capturing abrupt changes in the stability of context information flow.

Specifically, let $\mathcal{I}_c$ and $\mathcal{I}_q$ denote the token indices of the context examples $s_1$ and the query $s_2$, respectively.
For a perturbation applied at layer $l$, we intervene on the residual hidden states after the $l$-th Transformer block by masking only the context-token positions:
\begin{equation}
\tilde{h}_{l,i} =
\begin{cases}
\mathbf{0}, & i \in \mathcal{I}_c,\\
h_{l,i}, & i \in \mathcal{I}_q.
\end{cases}
\end{equation}

The perturbed hidden states $\tilde{H}_l$ are then fed into the remaining layers $l+1,\dots,N$ to obtain the final output distribution $Q(l)$.

Given the original intact output distribution $P$ and the perturbed distribution $Q(l)$ at layer $l$, we define the JS divergence at layer $l$ as:
\begin{equation}
\mathrm{JS}(l) = \mathrm{JS}\big(P \parallel Q(l)\big).
\end{equation}

The standard Jensen-Shannon divergence is formulated as:
\begin{equation}
\begin{aligned}
\mathrm{JS}\big(P \parallel Q(l)\big)
&= \frac{1}{2}\mathrm{KL}\big(P \parallel M(l)\big) \\
&+ \frac{1}{2}\mathrm{KL}\big(Q(l) \parallel M(l)\big),
\end{aligned}
\end{equation}
where $M(l) = \frac{1}{2}\big(P + Q(l)\big)$ represents the average mixed distribution, and $\mathrm{KL}(\cdot\parallel\cdot)$ denotes the Kullback-Leibler divergence:
\begin{equation}
\mathrm{KL}(X \parallel Y) = \sum_{i} X(i) \log \frac{X(i)}{Y(i)}.
\end{equation}

Based on the layer-level JS divergence sequence, we further compute the relative fluctuation between adjacent layers:
\begin{equation}
\Delta\mathrm{JS}(l) = \left| \frac{\mathrm{JS}(l) - \mathrm{JS}(l-1)}{\mathrm{JS}(l-1) + \epsilon} \right|,
\label{eq:js_definition}
\end{equation}
A large $\Delta\mathrm{JS}(l)$ value indicates an abrupt shift in layer sensitivity, meaning the stability of information flow changes significantly at this layer. This metric effectively identifies critical layers that are sensitive to parameter updates.

\section{Experiment}
\label{sec:experiment}
To empirically validate the diagnostic utility of LayerTracer and derive actionable freeze-train rules, we conduct comprehensive layer-wise analysis across Qwen3 variants. Figure~\ref{fig:main_results} illustrates the distribution of Task Particle and Layer-wise Sensitivity.
\subsection{Experimental Settings}
\noindent \textbf{Dataset.} We use a composite dataset constructed upon AntSynNET~\cite{nguyen:17antonym}, following four standard categories for layer-wise reasoning analysis: Knowledge–Algorithmic, Extractive–Knowledge, Extractive–Algorithmic, and Knowledge–Translation.
We reorganize each sample into a structured form \(t = s_1 \oplus s_2\).
This structural separation enables \(\boldsymbol{s_1}\)-targeted masking perturbation, which helps distinguish the roles of shallow and deep layers by isolating context influence during layer-wise sensitivity evaluation. For clarity, dataset details are presented in Appendix~\ref{sec:appendix_2}.

\noindent \textbf{Experimental Details.}
All experiments are conducted on a server equipped with 8 NVIDIA H200 141G GPUs.This study uses three Qwen3 variants of 0.6B, 8B, and 14B parameters, as the primary subjects for layer-wise analysis. The experiment aims to characterize the distribution trends of Task Particles and Layer-wise Sensitivity across varying model capacities. For each layer, the hidden state is projected via the language modeling head to obtain the full vocabulary distribution, from which we retain only the top-10 candidate tokens with the highest probabilities. We verify in Appendix~\ref{sec:appendix_3_1} that results remain consistent when extending to top-50, confirming the robustness of our layer-wise patterns.

\subsection{Experimental Results}
\subsubsection{Deep Layers as Execution Zones}

\vspace{0.5em}
{
\centering
\includegraphics[width=\columnwidth]{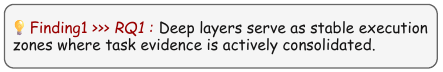}\par
}

Figure~\ref{fig:a}--\ref{fig:c} visualize the layer-wise distribution of $\mathrm{Ratio}(l)$, the relative change of task probability defined in Equation~\ref{eq:ratio_definition}. Following our definition in Section~\ref{sec:task_particle}, layers with $\mathrm{Ratio}(l) > 0$ constitute the Task Particle interval, marking where task evidence undergoes meaningful shifts. We observe that high $\mathrm{Ratio}$ values (darker green) concentrate in deep layers across all model scales, forming a stable task-execution zone.  This indicates that deep layers serve as the primary region where the model executes task-specific reasoning, while shallow layers show minimal task-related activity.

\begin{figure}[t]
    \centering
    \subfloat[Nemotron-12B-Base]{\label{fig:a2}
        \includegraphics[width=0.48\textwidth]{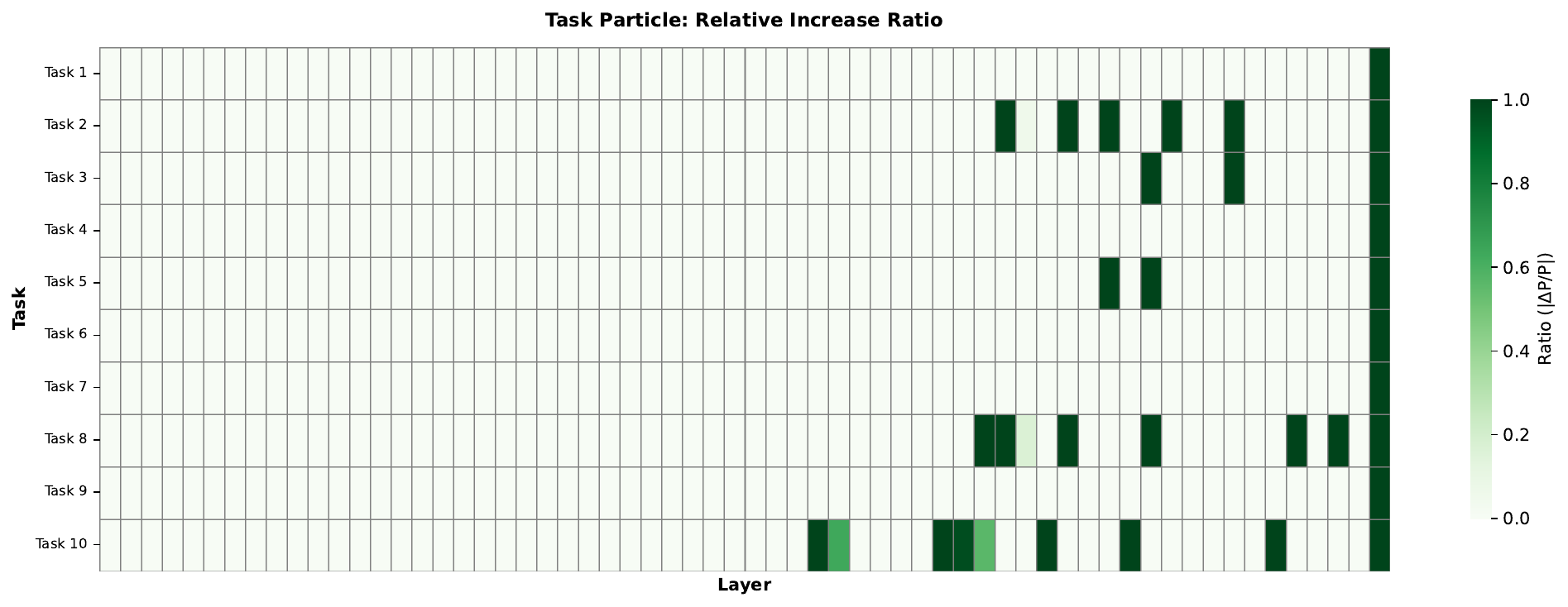}}

    \subfloat[Nemotron-12B-Base]{\label{fig:d2}
        \includegraphics[width=0.48\textwidth]{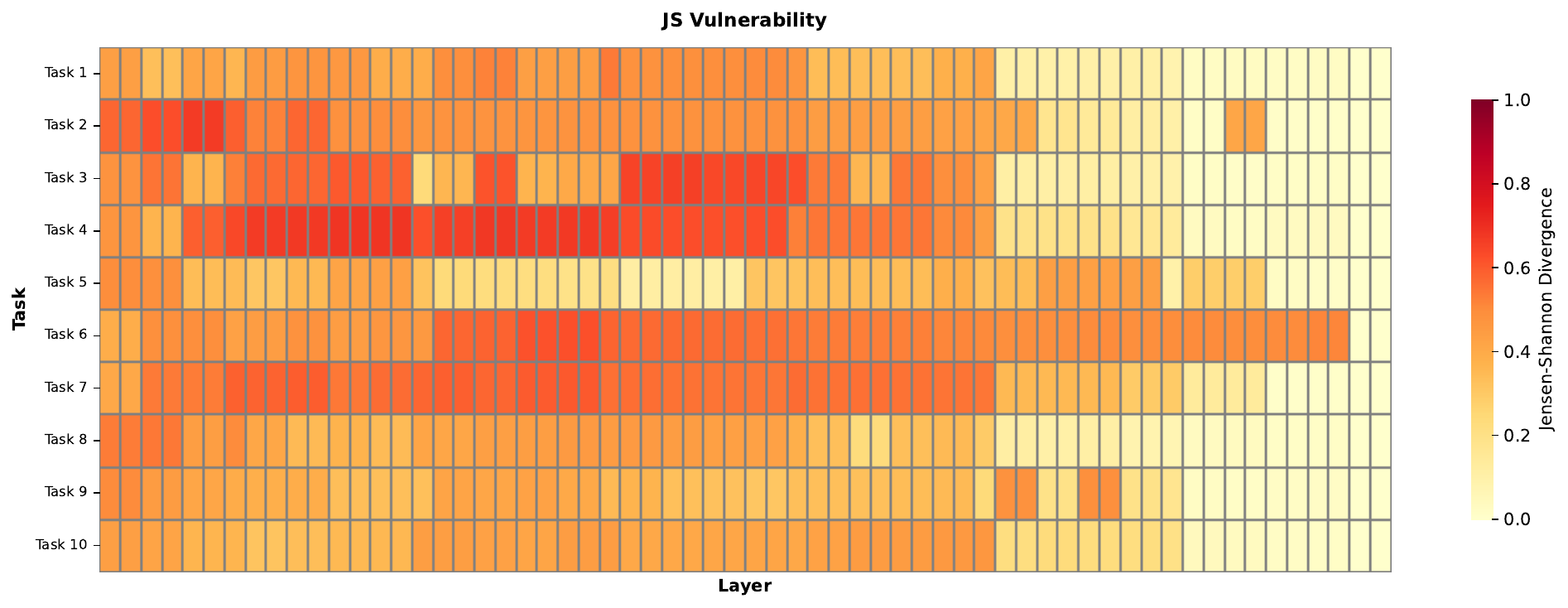}}
    \hfill
    \caption{Layer-wise analysis results on Nemotron-12B-Base, a Mamba-based non-Transformer architecture. Consistent with observations on the Qwen3 series, the Task Particle concentrates in deep layers while shallow layers exhibit higher instability, verifying the universality of our findings across diverse model architectures.
    }
    \label{fig:main_results2} 
\end{figure}

\subsubsection{Shallow Layers as Sensitive Zones}

\vspace{0.5em}
{
\centering
\includegraphics[width=\columnwidth]{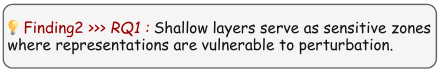}\par
}

Figure~\ref{fig:d}--\ref{fig:f} illustrate $\Delta\mathrm{JS}(l)$, which quantifies the 
relative change in layer-wise sensitivity under masking perturbations. Based on Equation~\ref{eq:js_definition}, higher $\Delta\mathrm{JS}$ values (darker red) indicate 
layers where sensitivity to context perturbations changes abruptly. The results reveal that shallow layers exhibit significantly larger $\Delta\mathrm{JS}$ fluctuations compared to deep layers, which maintain stable sensitivity profiles. This indicates that shallow layers remain highly responsive to parameter updates, making them natural candidates for adaptation, while deep layers have consolidated stable knowledge representations.

\begin{table*}[t]
\centering
\small
\setlength{\tabcolsep}{4pt}
\renewcommand{\arraystretch}{1.2} 
\begin{tabularx}{\textwidth}{l *{5}{>{\centering\arraybackslash}X}}
\toprule
\multirow{2.5}{*}{\textbf{Models}} & \multicolumn{5}{c}{\textbf{CEval}} \\
\cmidrule(lr){2-6}
& \textbf{STEM} & \textbf{Social Science} & \textbf{Humanities} & \textbf{Other} & \textbf{Average} \\
\midrule
Qwen3-0.6B-Base (Trainable) & 26.60 / 26.55 & 27.19 / 27.21 & 26.18 / 25.83 & 26.53 / 25.67 & 26.62 / 26.32 \\
\midrule
Qwen3-0.6B-Base (Frozen/Trainable) & \textcolor{green!60!black}{24.86} / \textcolor{green!60!black}{24.97} & \textcolor{green!60!black}{25.33} / \textcolor{green!60!black}{25.60} & \textcolor{green!60!black}{25.51} / \textcolor{green!60!black}{25.71} & \textcolor{green!60!black}{25.85} / \textcolor{red!60!black}{25.92} & \textcolor{green!60!black}{25.39} / \textcolor{green!60!black}{25.55} \\
\midrule
Qwen3-0.6B-Base (Trainable/Frozen) & \textcolor{red!60!black}{\textbf{27.76}} / \textcolor{red!60!black}{\textbf{27.65}} & \textcolor{red!60!black}{\textbf{32.07}} / \textcolor{red!60!black}{\textbf{32.17}} & \textcolor{red!60!black}{\textbf{28.94}} / \textcolor{red!60!black}{\textbf{29.39}} & \textcolor{red!60!black}{\textbf{28.82}} / \textcolor{red!60!black}{\textbf{28.99}} & \textcolor{red!60!black}{\textbf{29.40}} / \textcolor{red!60!black}{\textbf{29.55}} \\
\bottomrule
\end{tabularx}
\caption{Performance evaluation on the C-Eval benchmark~\cite{huang2023ceval} across three training strategies for Qwen3-0.6B-Base. Each cell reports Generate / Logit scores. The first row denotes the baseline with full-parameter pre-tuning. The second row represents the strategy of freezing shallow layers and training deep layers. The third row illustrates our strategy of training shallow layers and freezing deep layers.
\textcolor{red!60!black}{Red values} indicates performance better than the full-parameter baseline, while \textcolor{green!60!black}{green values} indicates performance worse than the baseline. \textbf{Bold values} denote the best performance across all settings. Standard deviation details over five repeated runs for all experiments are provided in Appendix~\ref{sec:zero_std}.
} 
\label{tab:ceval_performance}
\end{table*}

\begin{table*}[t]
\centering
\small
\setlength{\tabcolsep}{4pt}
\renewcommand{\arraystretch}{1.2} 
\begin{tabularx}{\textwidth}{l *{5}{>{\centering\arraybackslash}X}}
\toprule
\multirow{2.5}{*}{\textbf{Models}} & \multicolumn{5}{c}{\textbf{CMMLU}} \\
\cmidrule(lr){2-6}
& \textbf{STEM} & \textbf{Social Science} & \textbf{Humanities} & \textbf{Other} & \textbf{Average} \\
\midrule
Qwen3-0.6B-Base (Trainable) & 25.22 / 24.99 & 24.10 / 25.00 & 25.26 / 25.53 & 25.85 / 25.54 & 25.11 / 25.27 \\
\midrule
Qwen3-0.6B-Base (Frozen/Trainable) & \textcolor{green!60!black}{24.28} / \textcolor{green!60!black}{24.14} & \textcolor{red!60!black}{24.98} / \textcolor{red!60!black}{26.01} & \textcolor{green!60!black}{25.14} / \textcolor{green!60!black}{25.50} & \textcolor{red!60!black}{26.30} / \textcolor{red!60!black}{26.80} & \textcolor{red!60!black}{25.17} / \textcolor{red!60!black}{25.61} \\
\midrule
Qwen3-0.6B-Base (Trainable/Frozen) & \textcolor{red!60!black}{\textbf{26.69}} / \textcolor{red!60!black}{\textbf{26.72}} & \textcolor{red!60!black}{\textbf{30.62}} / \textcolor{red!60!black}{\textbf{30.65}} & \textcolor{red!60!black}{\textbf{27.93}} / \textcolor{red!60!black}{\textbf{28.00}} & \textcolor{red!60!black}{\textbf{27.76}} / \textcolor{red!60!black}{\textbf{28.32}} & \textcolor{red!60!black}{\textbf{28.25}} / \textcolor{red!60!black}{\textbf{28.42}} \\
\bottomrule
\end{tabularx}
\caption{Performance evaluation on the CMMLU benchmark~\cite{li-etal-2024-cmmlu} across the same three pre-training strategies. Settings and notations follow Table~\ref{tab:ceval_performance}.}
\label{tab:cmmlu_performance}
\end{table*}

\subsubsection{Core Principle: Matching Zones with Strategies}
\label{sec:Principle}

\vspace{0.5em}
{
\centering
\includegraphics[width=\columnwidth]{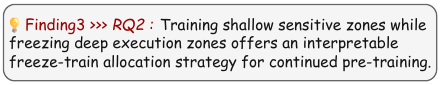}\par
}

The above results reveal a consistent shallow-sensitive and deep-stable hierarchy, suggesting a simple layer-wise allocation principle for continued pre-training. We instantiate this principle with a model-midpoint split: the shallow half is treated as sensitive zones, where high $\Delta\mathrm{JS}$ indicates greater sensitivity to perturbations and thus higher plasticity for absorbing domain-specific changes. In contrast, the deep half is treated as execution zones, where high $\mathrm{Ratio}$ values indicate active task-evidence consolidation and stabilized task execution. Since these deep layers carry task-execution functions and inherit high-quality parameters from strong pre-trained models, they should be preserved rather than frequently updated. This midpoint split provides a symmetric parameter partition and a clean controlled setting for validation, instead of exhaustively searching for the optimal boundary.

\begin{figure}[t]
    \centering
    \includegraphics[width=\linewidth]{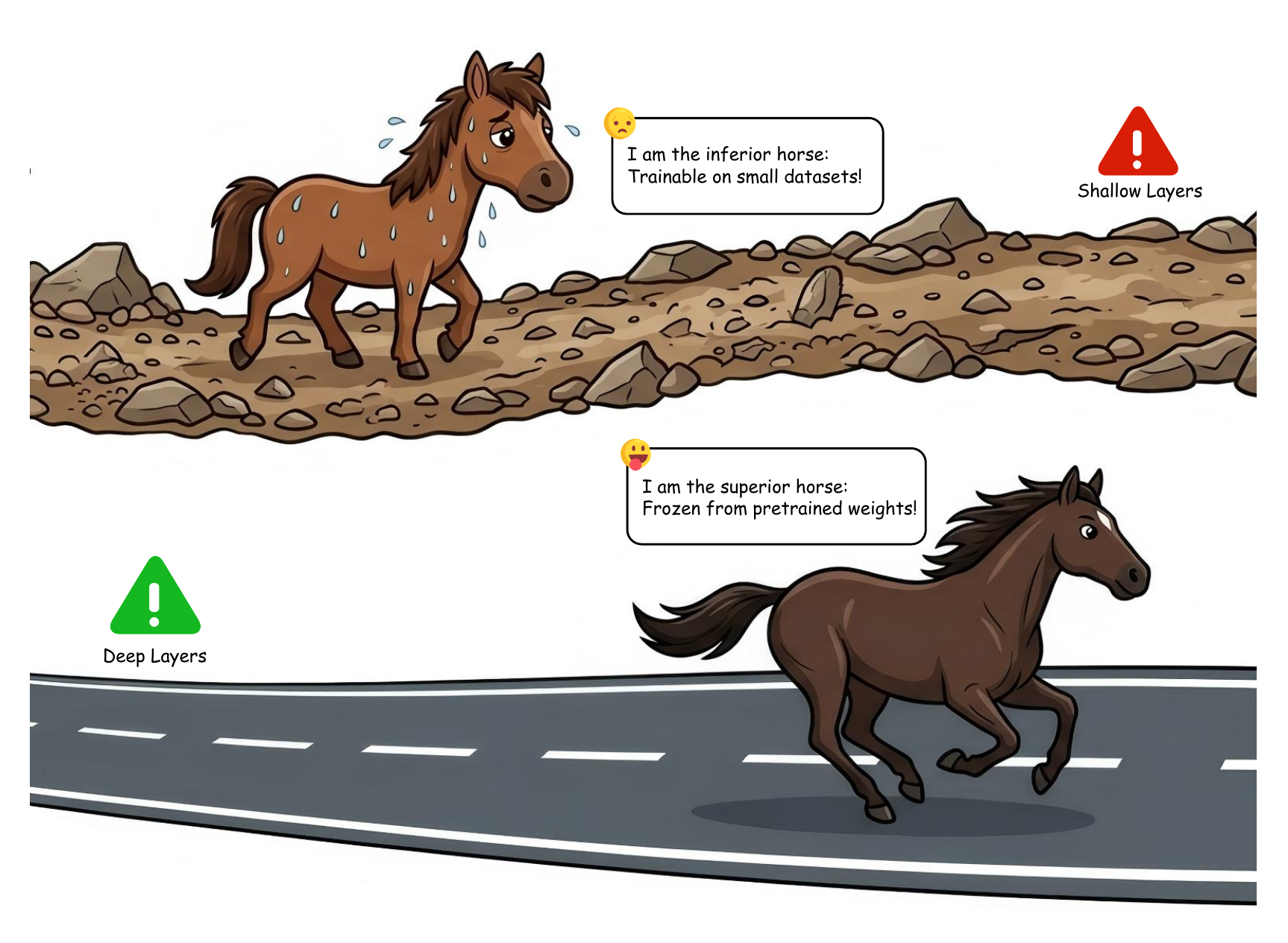}
    \caption{Illustration of the proposed zone-strategy matching principle. Shallow layers are treated as sensitive zones, where active training supports domain-specific updates, while deep layers are treated as execution zones, where freezing preserves pre-trained task evidence.}
    \label{fig:exp_1}
\end{figure}

As illustrated in Figure~\ref{fig:exp_1}, we therefore match different strategies to different layer zones: shallow sensitive zones are trained to absorb domain-specific patterns, while deep execution zones are frozen to preserve strong pre-trained parameters and avoid disrupting consolidated task evidence. Details as elaborated in~\ref{sec:boundary_rationale}.

\section{Why is LayerTracer Effective?}
\label{sec:effective}

Guided by the hierarchical patterns revealed by LayerTracer in Section~\ref{sec:experiment}, we further evaluate whether the derived strategy leads to measurable gains in practical continued pre-training settings. While Section~\ref{sec:experiment} focuses on representation dynamics and intrinsic layer properties, this section examines whether the LayerTracer-guided allocation strategy improves downstream task performance.

\vspace{0.4em}
{
\centering
\includegraphics[width=\columnwidth]{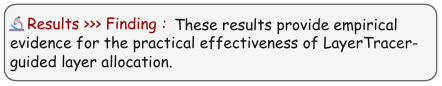}\par
}

\subsection{Pre-Training and Evaluation Setup}
\label{sec:train_setting}
We conduct continued pre-training on Qwen3-0.6B-Base using the high-quality Chinese corpus CCI3.0-HQ~\cite{wang2024cci30hqlargescalechinesedataset} with 262B tokens, where all training samples are in the form of natural text streams rather than question–answer pairs, and compare three strategies: full-parameter pre-training as the baseline, freezing the shallow module while training the deep module, and training the shallow module while freezing the deep module. Evaluation is performed on two standard Chinese benchmarks, C-Eval~\cite{huang2023ceval} and CMMLU~\cite{li-etal-2024-cmmlu}, using two mainstream protocols widely adopted for pre-trained model assessment: logit-based evaluation infers answers directly from model logits to reflect inherent knowledge fitting, while generation-based evaluation judges whether the correct answer is included within a 16-token autoregressive output with \texttt{do\_sample=False} to characterize generation quality, and consistency between these two metrics indicates reliable model behavior. Detailed settings are summarized in Appendix~\ref{sec:appendix_1}.

\begin{table}[t]
\centering
\small
\setlength{\tabcolsep}{16pt}
\begin{tabular}{lc}
\toprule
\textbf{Dataset} & \textbf{Tokens} \\
\midrule
Qwen Private Corpus & 36T \\
CCI3.0-HQ & 262B \\
\bottomrule
\end{tabular}
\caption{Statistics of pre-training corpora. The Qwen private corpus is a large-scale high-quality proprietary dataset built by the Qwen team, while CCI3.0-HQ~\cite{wang2024cci30hqlargescalechinesedataset} is a public high-quality Chinese corpus with a smaller scale.}
\label{tab:pretrain_corpus}
\end{table}
\subsection{Results and Analysis}
As shown in Table~\ref{tab:ceval_performance} and Table~\ref{tab:cmmlu_performance}, the train-shallow/freeze-deep strategy consistently achieves the best performance across both benchmarks and both evaluation protocols. On C-Eval, it improves the average score from 26.62\% / 26.32\% under full-parameter continued pre-training to 29.40\% / 29.55\%, yielding absolute gains of 2.78\% in Generate evaluation and 3.23\% in Logit evaluation. In contrast, the reverse strategy, which freezes shallow layers and trains deep layers, drops to 25.39\% / 25.55\%, falling below the full-parameter baseline.

A similar trend is observed on CMMLU. The train-shallow/freeze-deep strategy reaches 28.25\% / 28.42\% on average, outperforming the full-parameter baseline by 3.14\% and 3.15\% under Generate and Logit evaluation, respectively. By comparison, the freeze-shallow/train-deep strategy obtains only 25.17\% / 25.61\%, again showing clear degradation. These results indicate that updating deep layers can disrupt pre-trained task evidence, while preserving deep execution layers and adapting shallow layers leads to more reliable continued pre-training. More case studies are provied in Appendix~\ref{sec:appendix_4}.

\section{A Case Study: Industrial Implications}
\label{sec:case_study}

\vspace{0.4em}
{
\centering
\includegraphics[width=\columnwidth]{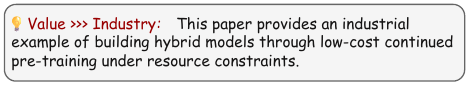}\par
}

\begin{table*}[t]
\centering
\small
\setlength{\tabcolsep}{4pt}
\renewcommand{\arraystretch}{1.2} 
\begin{tabularx}{\textwidth}{l *{5}{>{\centering\arraybackslash}X}}
\toprule
\cmidrule(lr){2-6}
& \textbf{STEM} & \textbf{Social Science} & \textbf{Humanities} & \textbf{Other} & \textbf{Average} \\
\midrule
Qwen3-Nemotron-Base & 17.39 / 24.78 & 17.95 / 24.65 & 14.65 / 25.60 & 15.79 / 26.94 & 16.45 / 25.49 \\
\midrule
Nemotron-Qwen3-Base & \textbf{20.40} / \textbf{29.40} & \textbf{19.77} / \textbf{34.30} & \textbf{20.37} / \textbf{26.78} & \textbf{20.19} / \textbf{28.48} & \textbf{20.19} / \textbf{29.73} \\
\bottomrule
\end{tabularx}
\caption{Performance evaluation on the C-Eval benchmark for two hybrid architectures. Nemotron-Qwen3-Base places pre-trained Qwen3 decoder layers in the deep half, while Qwen3-Nemotron-Base places them in the shallow half. Bold values denote the best performance.}
\label{tab:ceval_hybrid_performance}
\end{table*}

\begin{table*}[t]
\centering
\small
\setlength{\tabcolsep}{4pt}
\renewcommand{\arraystretch}{1.2} 
\begin{tabularx}{\textwidth}{l *{5}{>{\centering\arraybackslash}X}}
\toprule
\multirow{2.5}{*}{\textbf{Models}} & \multicolumn{5}{c}{\textbf{CMMLU}} \\
\cmidrule(lr){2-6}
& \textbf{STEM} & \textbf{Social Science} & \textbf{Humanities} & \textbf{Other} & \textbf{Average} \\
\midrule
Qwen3-Nemotron-Base & 17.25 / 24.59 & 18.73 / 24.62 & 15.75 / 24.66 & 15.83 / 24.67 & 16.89 / 24.64 \\
\midrule
Nemotron-Qwen3-Base & \textbf{20.79} / \textbf{26.87} & \textbf{21.76} / \textbf{29.94} & \textbf{22.21} / \textbf{27.95} & \textbf{19.43} / \textbf{28.43} & \textbf{20.80} / \textbf{28.30} \\
\bottomrule
\end{tabularx}
\caption{Performance evaluation on the CMMLU benchmark for the same two hybrid strategies. Settings and notations follow Table~\ref{tab:ceval_hybrid_performance}.}
\label{tab:cmmlu_hybrid_performance}
\end{table*}

\begin{figure}[!t]
    \centering
    \includegraphics[width=\columnwidth]{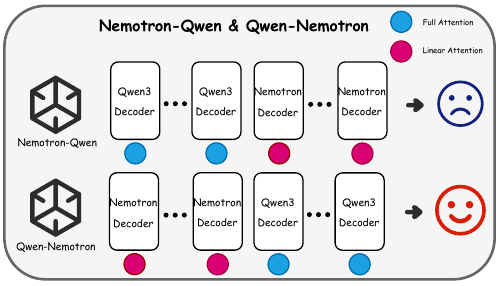}
    \caption{The architectures of Nemotron-Qwen model and Qwen-Nemotron model. They are hybrid architectures with a 1:1 ratio of Full Attention to Linear Attention.}
    \label{fig:hybrid_architecture}
\end{figure}

This section examines the industrial value of LayerTracer for resource-constrained LLM development. As hybrid architectures become more common, a key challenge is how to place heterogeneous modules while preserving the generalization ability of foundation models. Guided by the LayerTracer principle of preserving deep execution zones while adapting shallow sensitive zones, we provide a practical case study for hybrid model construction. Figure~\ref{fig:hybrid_architecture} illustrates the hybrid architecture used in this study.

To illustrate its industrial value, we construct two hybrid models, Nemotron-Qwen3-Base and Qwen3-Nemotron-Base, both with 550M parameters. In Nemotron-Qwen3-Base, Mamba2-based Nemotron layers are placed in the first half of the network and actively trained, while Qwen3 pre-trained decoder layers are placed in the second half and frozen to preserve high-quality pre-trained parameters. In Qwen3-Nemotron-Base, the allocation is reversed, placing Qwen3 layers in the first half and Nemotron layers in the second half. We adopt the same pre-training data, benchmarks, and evaluation metrics as Section~\ref{sec:effective}.

Experimental results are reported in Table~\ref{tab:ceval_hybrid_performance} and Table~\ref{tab:cmmlu_hybrid_performance}. On C-Eval, placing Qwen3 pre-trained layers in the deep half improves the average score from 16.45\% / 25.49\% to 20.19\% / 29.73\%, yielding absolute gains of 3.74 and 4.24 percentage points under Generate and Logit evaluation, respectively. On CMMLU, the same strategy improves the average score from 16.89\% / 24.64\% to 20.80\% / 28.30\%, with gains of 3.91 and 3.66 percentage points. The improvements are consistent across STEM, Social Science, Humanities, and Other categories.

\section{Conclusion}
This paper proposes LayerTracer, an interpretable diagnostic framework for layer-wise freeze-train allocation in continued pre-training. By combining Task Particle and Layer-wise Sensitivity, LayerTracer reveals a consistent hierarchy: shallow layers are more sensitive and suitable for adaptation, while deep layers form stable execution zones where task evidence is consolidated. Based on this finding, we derive a train-shallow/freeze-deep strategy. Experiments show that this strategy outperforms both full-parameter continued pre-training and the reverse allocation, while the hybrid architecture case study further demonstrates the value of preserving high-quality pre-trained parameters in deep layers. Overall, LayerTracer provides practical and low-cost guidance for efficient model adaptation and hybrid model construction.

\section*{Limitations}

LayerTracer is designed as an interpretable diagnostic framework for continued pre-training, and its main limitation is computational scale rather than methodological scope. Due to limited resources, we validate it on mainstream model sizes instead of extending the analysis to much larger parameter scales. Even so, our experiments already cover comparatively large continued pre-training settings, and the results are fully reproducible, stable, and consistent across benchmarks. We believe the framework provides practical and effective guidance for layer-wise allocation, while larger-scale validation remains an important direction for future work.

\section*{Ethical Statement}

This work focuses on fundamental interpretability and efficient adaptation techniques for large language models, with no involvement of unethical data collection, harmful content generation, or biased application scenarios. All experiments comply with standard AI research ethics and model usage norms. The proposed LayerTracer framework aims to improve training efficiency and stability while reducing computational costs, which supports green and responsible development of large models.

\section*{Acknowledgments}
The research work described in this paper has been supported by the National Key R\&D Program of China (No. 2025YFG0100800). The authors would like to thank the anonymous reviewers for their valuable comments and suggestions to improve this paper.



\bibliography{custom}

\clearpage 
\appendix

\section{Training and Evaluation Details}
\label{sec:appendix_1}

To ensure rigorous and reproducible comparisons, all continued pre-training experiments were conducted under a unified configuration. This appendix details the hardware setup, hyperparameter settings, and training efficiency metrics across all evaluated layer-wise allocation strategies.

\subsection{Hardware and Implementation.}
All experiments were implemented using the native Accelerate training framework and executed on a single server equipped with 8 $\times$ NVIDIA H200 141G. To strictly isolate the impact of layer-wise parameter allocation, we maintained identical data pipelines, hardware environments, and training schedules across all trials. Wall-clock training time was measured from the first optimizer step to the final checkpoint save, explicitly excluding data loading and model initialization overhead.
\begin{table}[!t]
\centering
\small
\begin{tabular}{ll}
\toprule
\textbf{Hyperparameter} & \textbf{Value} \\
\midrule
Sequence Length & 4096 \\
Precision & BF16 \\
Optimizer (AdamW) & $\beta_1=0.9$, $\beta_2=0.95$ \\
Learning Rate & $3 \times 10^{-5}$ \\
Warmup Ratio & 0.1 \\
Weight Decay & 0.01 \\
Per-GPU Micro Batch Size & 8 \\
Gradient Accumulation Steps & 1 \\
Global Batch Size & 64 \\
Training Epochs & 1 \\
Training Tokens & \textasciitilde 262B \\
\bottomrule
\end{tabular}
\caption{Unified hyperparameter configuration for all continued pre-training experiments.}
\label{tab:hyperparams}
\end{table}

\subsection{Hyperparameter Configuration}
We adopted a consistent hyperparameter configuration for all continued pre-training runs, as summarized in Table~\ref{tab:hyperparams}. Models were trained on the CCI3.0-HQ corpus (\textasciitilde 262B tokens) for a single epoch with a sequence length of 4096 and BF16 precision. Optimization was performed using AdamW ($\beta_1=0.9$, $\beta_2=0.95$) with a learning rate of $3 \times 10^{-5}$, a linear warmup ratio of 0.1, and a weight decay of 0.01. The effective global batch size of 64 was achieved using a per-GPU micro-batch size of 8 with a single gradient accumulation step.

\subsection{Training Efficiency and Strategy Mapping.}
Table~\ref{tab:efficiency} reports the computational efficiency across all evaluated strategies. Strategies (1)--(3) correspond to the vanilla Qwen3-0.6B-Base variants reported in Table~\ref{tab:ceval_performance} of the main text: (1) full-parameter training, (2) Train-Shallow/Freeze-Deep, and (3) Freeze-Shallow/Train-Deep. Strategies (4) and (5) correspond to the hybrid architectures evaluated in Table~\ref{tab:ceval_hybrid_performance}: (4) Nemotron-Qwen3-Base (training shallow NemotronDecoderLayer while freezing deep QwenDecoderLayer) and (5) Qwen3-Nemotron-Base (training shallow QwenDecoderLayer while freezing deep NemotronDecoderLayer).

As shown in Table~\ref{tab:efficiency}, partial-parameter strategies substantially reduce the number of trainable parameters, from approximately 0.6B in full-parameter training to about 0.3B in the vanilla partial-training setting and 0.23B in the hybrid setting. However, the wall-clock training time does not decrease monotonically with the number of trainable parameters. For the vanilla Qwen3-0.6B-Base variants, both partial-training strategies reduce training time compared with full-parameter training, from approximately 141h to 112h and 107h, respectively. In contrast, the hybrid architectures introduce additional optimization difficulty due to heterogeneous layer composition. In particular, when trainable modules must adapt their representations to frozen modules from a different architecture family, the optimization process involves not only parameter updating but also cross-architecture representation alignment. This explains why the hybrid strategies can require comparable or even longer wall-clock time despite involving fewer trainable parameters.

\begin{figure*}[t]
    \centering
    \subfloat[Train-Shallow/Freeze-Deep]{\label{fig:a3}
        \includegraphics[width=0.32\textwidth]{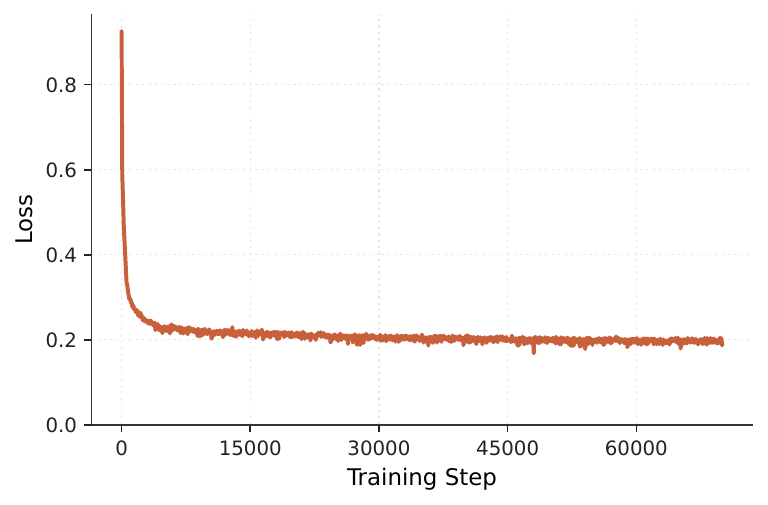}}
    \hfill
    \subfloat[Freeze-Shallow/Train-Deep]{\label{fig:b3}
        \includegraphics[width=0.32\textwidth]{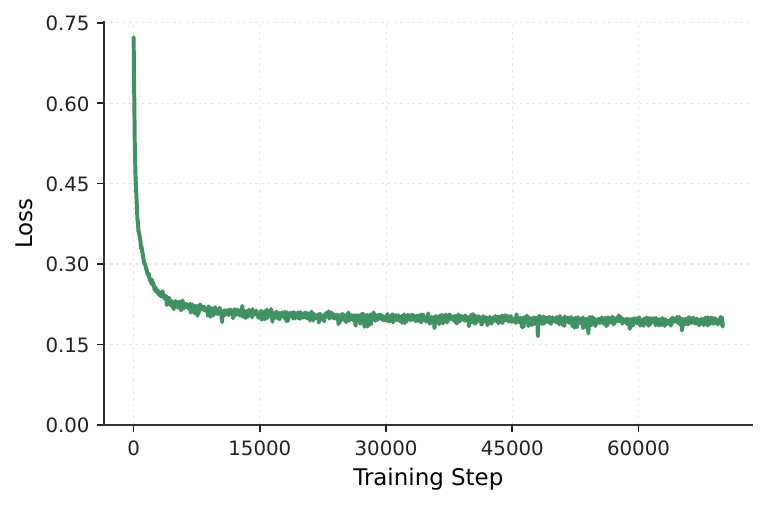}}
    \hfill
    \subfloat[Full-parameter Training]{\label{fig:c3}
        \includegraphics[width=0.32\textwidth]{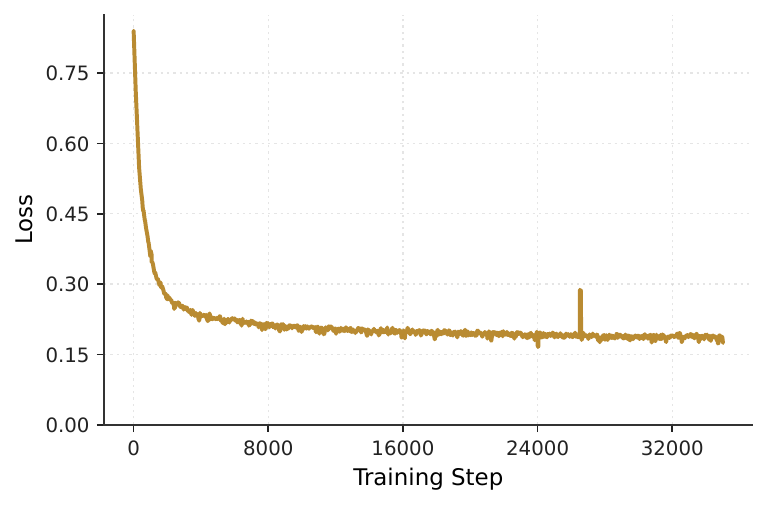}}
    
    \caption{Training loss curves across three layer-wise allocation strategies on the CCI3.0-HQ corpus. 
    (a) Train-Shallow/Freeze-Deep: shallow layers are updated while deep layers are frozen, showing stable convergence and the lowest final loss. 
    (b) Freeze-Shallow/Train-Deep: deep layers are updated while shallow layers are frozen, exhibiting slower convergence and higher final loss due to disruption of task-critical representations. 
    (c) Full-parameter Training: all layers are updated, serving as the baseline with intermediate convergence behavior. 
    The consistent pattern validates that preserving deep execution zones while adapting shallow sensitive zones yields more efficient and stable continued pre-training.}
    \label{fig:loss} 
\end{figure*}

Figure~\ref{fig:loss} illustrates the training loss curves of the three models investigated in Section~\ref{sec:experiment}. Overall, the three curves show broadly similar decreasing trends under the same corpus and training schedule, indicating stable continued pre-training across different allocation strategies. 

\subsection{Evaluation Details}
To ensure fair comparison with existing baselines and official leaderboards, all evaluations are conducted in a zero-shot setting without any task-specific demonstrations or fine-tuning. The evaluation prompts strictly follow the official templates provided by C-Eval~\cite{huang2023ceval} and CMMLU~\cite{li-etal-2024-cmmlu}, preserving the original question formatting, option ordering, and answer extraction rules. For generation-based evaluation, we adopt the official decoding configuration with temperature set to $0.95$, top-p sampling disabled, and a maximum output length of $16$ tokens to align with the benchmark protocols. For logit-based evaluation, we directly extract the probabilities of candidate answer tokens from the model's vocabulary logits without any decoding,. All evaluations are performed on the validation sets of both benchmarks, and results are reported as accuracy percentages across four domain categories (STEM, Social Science, Humanities, and Other) as well as the overall average.

\begin{table*}[!t]
\centering
\small
\renewcommand{\arraystretch}{1.3}
\begin{tabular}{lcc}
\toprule
\textbf{Strategy} & \textbf{Trainable Parameters} & \textbf{Training Time} \\
\midrule
(1) Full-Parameter & \textasciitilde 0.6B & \textasciitilde 141h \\
(2) Train-Shallow / Freeze-Deep & \textasciitilde 0.3B & \textasciitilde 107h \\
(3) Train-Deep / Freeze-Shallow & \textasciitilde 0.3B & \textasciitilde 112h \\
(4) Hybrid Architectures (Train-Shallow) & \textasciitilde 0.23B & \textasciitilde 115h \\
(5) Hybrid Architectures (Train-Deep) & \textasciitilde 0.23B & \textasciitilde 146h \\
\bottomrule
\end{tabular}
\caption{Training efficiency comparison across layer-wise allocation strategies. Strategies (1)-(3) correspond to vanilla Qwen3-0.6B-Base variants, while (4)-(5) correspond to the hybrid architectures (Nemotron-Qwen3-550M-Base and Qwen3-Nemotron-550M-Base).}
\label{tab:efficiency}
\end{table*}

\begin{table*}[t]
\centering
\small
\setlength{\tabcolsep}{16pt} 
\begin{tabular}{ll}
\hline
\textbf{ID} & \textbf{Prompt} \\
\hline
1  & \texttt{Example:good->Bad, no-Yes; Query:bad->} \\
2  & \texttt{Example:hot->Cold, big-Small; Query:cold->} \\
3  & \texttt{Example:fast->Slow, light-Heavy; Query:slow->} \\
4  & \texttt{Example:love->Hate, start-End; Query:hate->} \\
5  & \texttt{Example:day->Night, up-Down; Query:night->} \\
6  & \texttt{Example:rich->Poor, win-Lose; Query:poor->} \\
7  & \texttt{Example:early->Late, high-Low; Query:late->} \\
8  & \texttt{Example:strong->Weak, loud-Quiet; Query:weak->} \\
9  & \texttt{Example:hard->Soft, dark-Bright; Query:soft->} \\
10 & \texttt{Example:full->Empty, open-Close; Query:empty->} \\
\hline
\end{tabular}
\caption{Representative structured prompts from the AntSynNET-based dataset. These examples illustrate the prompt format rather than the full set of visualization groups.}
\label{tab:antsynnet_samples}
\end{table*}

\section{Experimental Dataset}
\label{sec:appendix_2}

\subsection{Dataset Construction}
\label{sec:appendix_2_1}
The original AntSynNET dataset~\citep{nguyen:17antonym} provides word pairs for lexical semantic relation tasks, where each instance is represented as a tuple
$\langle w_1, w_2 \rangle$ indicating an antonym or synonym relation, such as \texttt{(good, bad)} or \texttt{(hot, cold)}.
To support our analysis in Section~\ref{sec:experiment}, we reformulate each instance into a structured prompt
\(t = s_1 \oplus s_2\), where \(s_1\) contains contextual demonstration pairs and \(s_2\) contains the query word to be predicted. The structured prompt follows the template:
\begin{center}
\small
\begin{tabular}{l}
\texttt{Example: [pair\_1], [pair\_2];} \\
\texttt{Query: [target\_word]->}
\end{tabular}
\end{center}

Here, $s_1$ spans the contextual examples before the query, while $s_2$ corresponds to the query segment.
Specifically, we perturb token positions belonging to $s_1$ while keeping the query segment $s_2$ unchanged.
In this way, the task query itself remains intact, and the resulting distributional change mainly reflects how each layer responds to contextual information disruption.
Thus, $s_1$-targeted masking serves as a unified and controlled disturbance for measuring layer-wise sensitivity.

\subsection{Why Controlled Lexical Prompts}
\label{sec:appendix_2_2}
LayerTracer tracks how the probability of a target token evolves across layers by projecting intermediate hidden states through the LM head.
Therefore, the diagnostic setting requires the first predicted token to be meaningful and directly related to the answer.
Complex QA benchmarks, such as math or medical QA, are less suitable for this purpose because the model may need to generate a reasoning process or follow a specific answer format. In contrast, the structured AntSynNET prompts provide short and controlled one-step lexical prediction tasks, making it easier to isolate how contextual demonstrations affect target-token emergence across layers.

Our analysis uses 500 structured samples in total.
For visualization, these samples are evenly divided into 10 groups, with 50 samples per group, denoted as Task~1--Task~10.
Layer-wise metrics are computed for each sample and then averaged within each group to produce the heatmap rows.
Table~\ref{tab:antsynnet_samples} provides 10 representative prompts to illustrate the dataset format.

\begin{figure*}[t]
    \centering
    \subfloat[Qwen3-0.6B-Base]{\label{fig:top50_ratio_a}
        \includegraphics[width=0.32\textwidth]{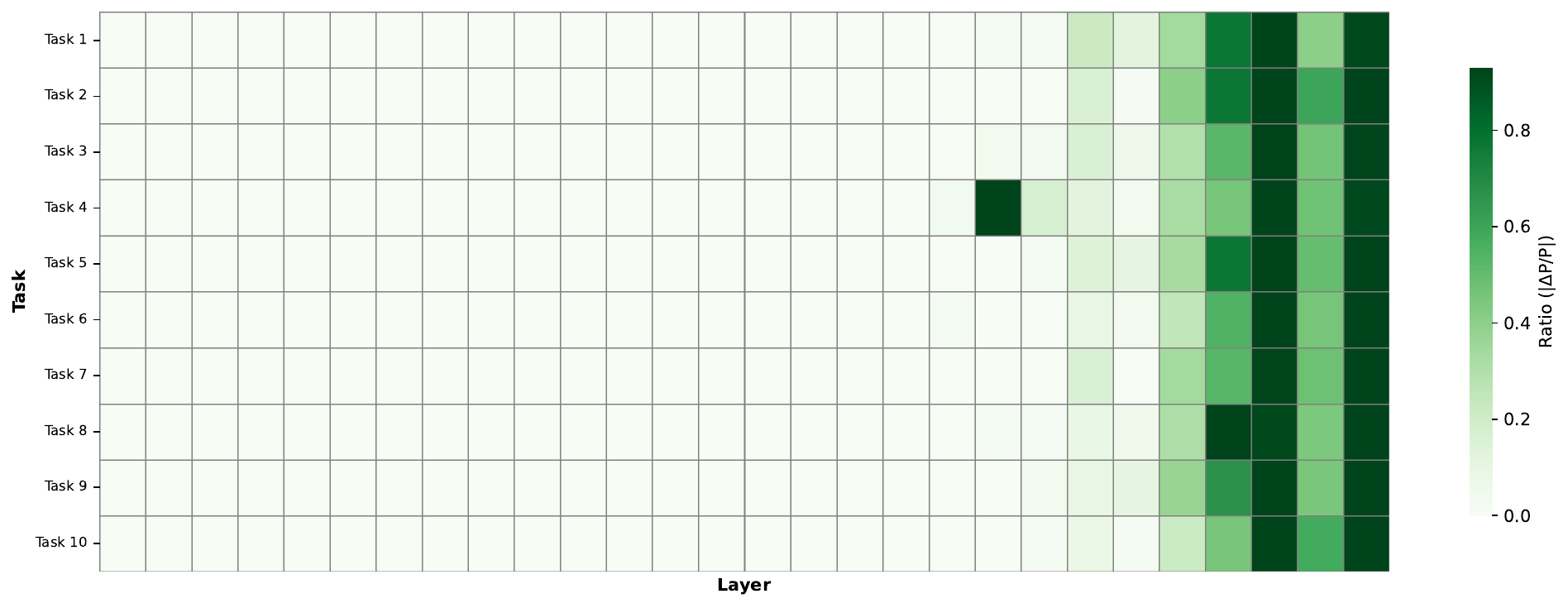}}
    \hfill
    \subfloat[Qwen3-8B-Base]{\label{fig:top50_ratio_b}
        \includegraphics[width=0.32\textwidth]{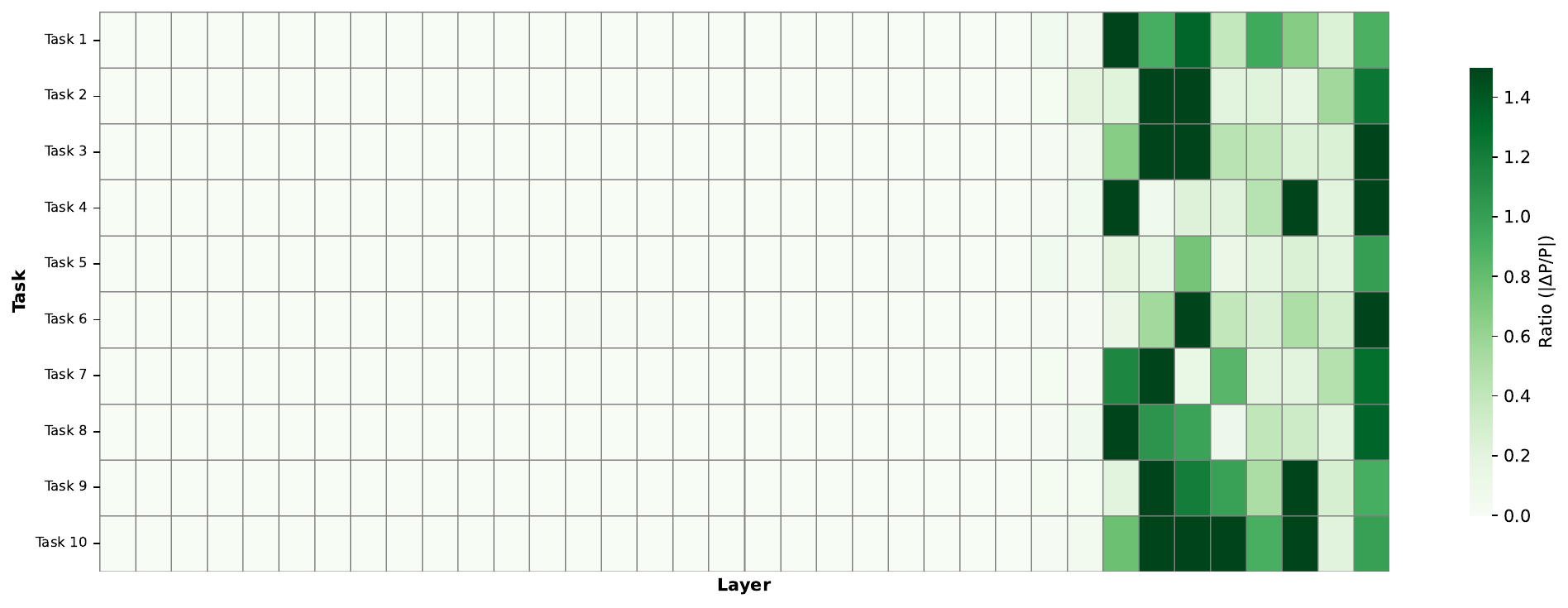}}
    \hfill
    \subfloat[Qwen3-14B-Base]{\label{fig:top50_ratio_c}
        \includegraphics[width=0.32\textwidth]{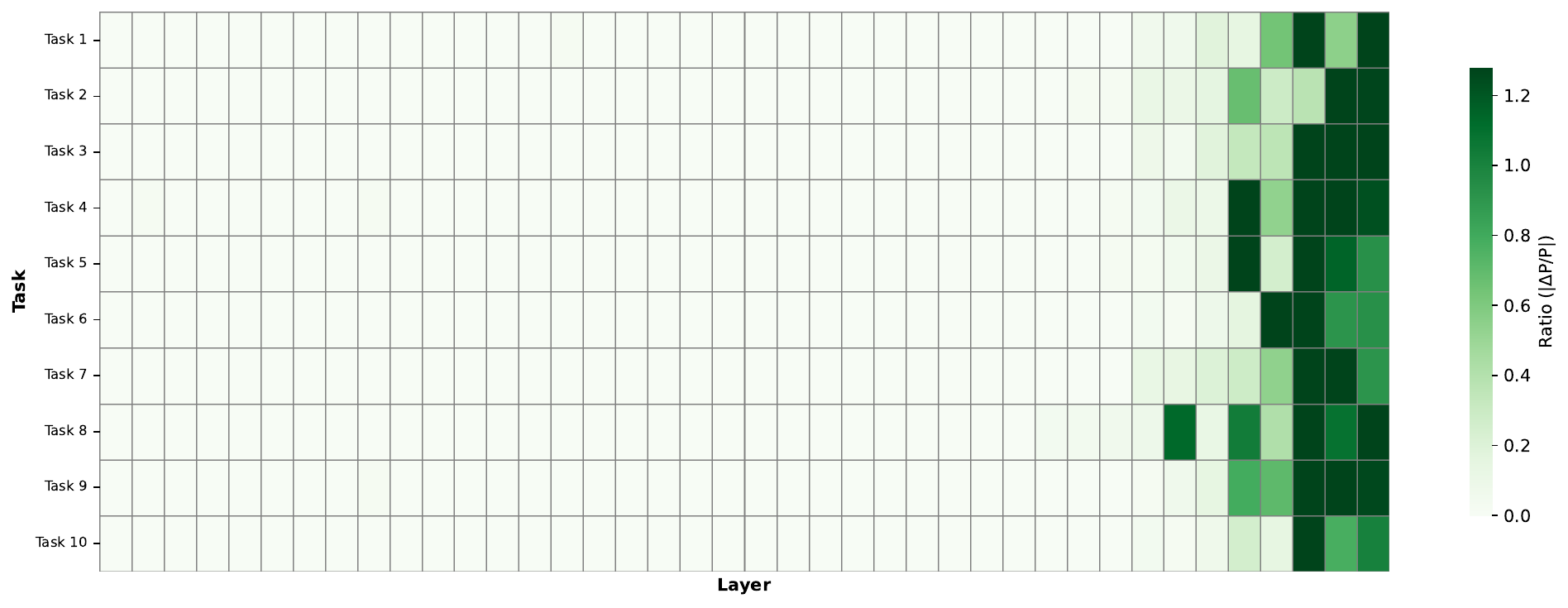}}

    \vspace{0.5em}

    \subfloat[Qwen3-0.6B-Base]{\label{fig:top50_js_d}
        \includegraphics[width=0.32\textwidth]{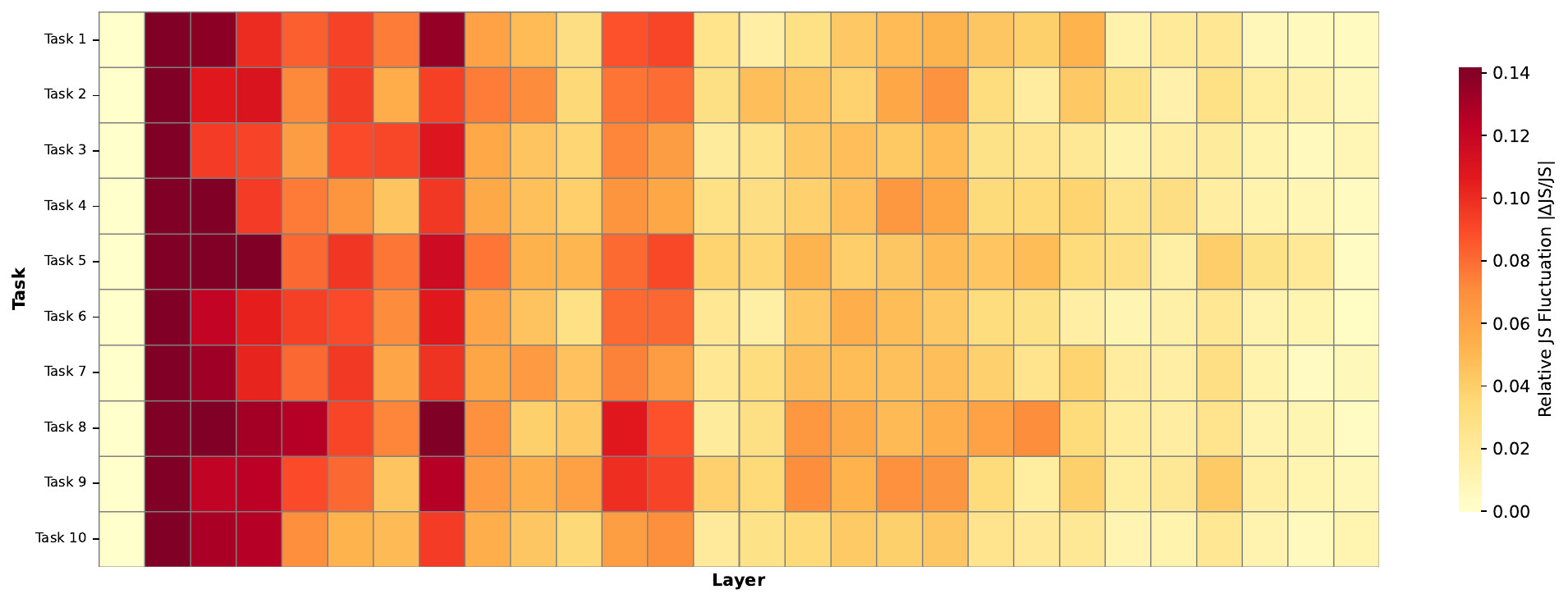}}
    \hfill
    \subfloat[Qwen3-8B-Base]{\label{fig:top50_js_e}
        \includegraphics[width=0.32\textwidth]{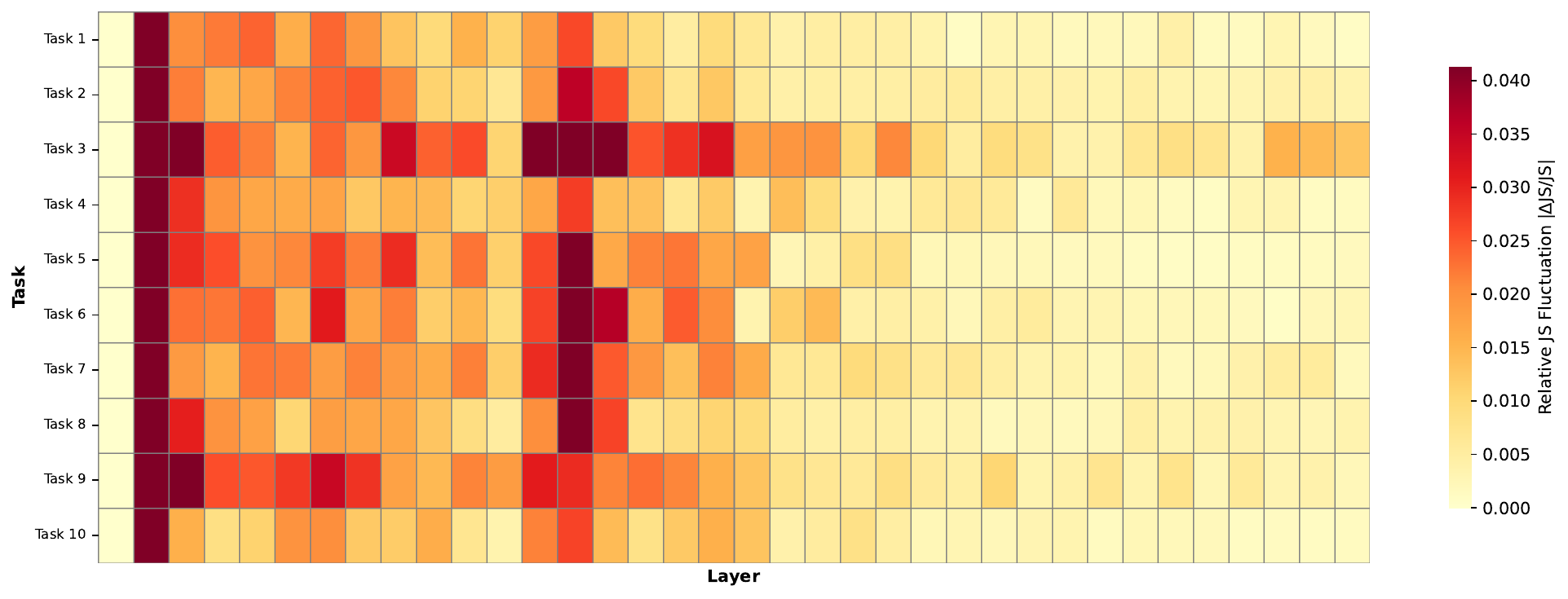}}
    \hfill
    \subfloat[Qwen3-14B-Base]{\label{fig:top50_js_f}
        \includegraphics[width=0.32\textwidth]{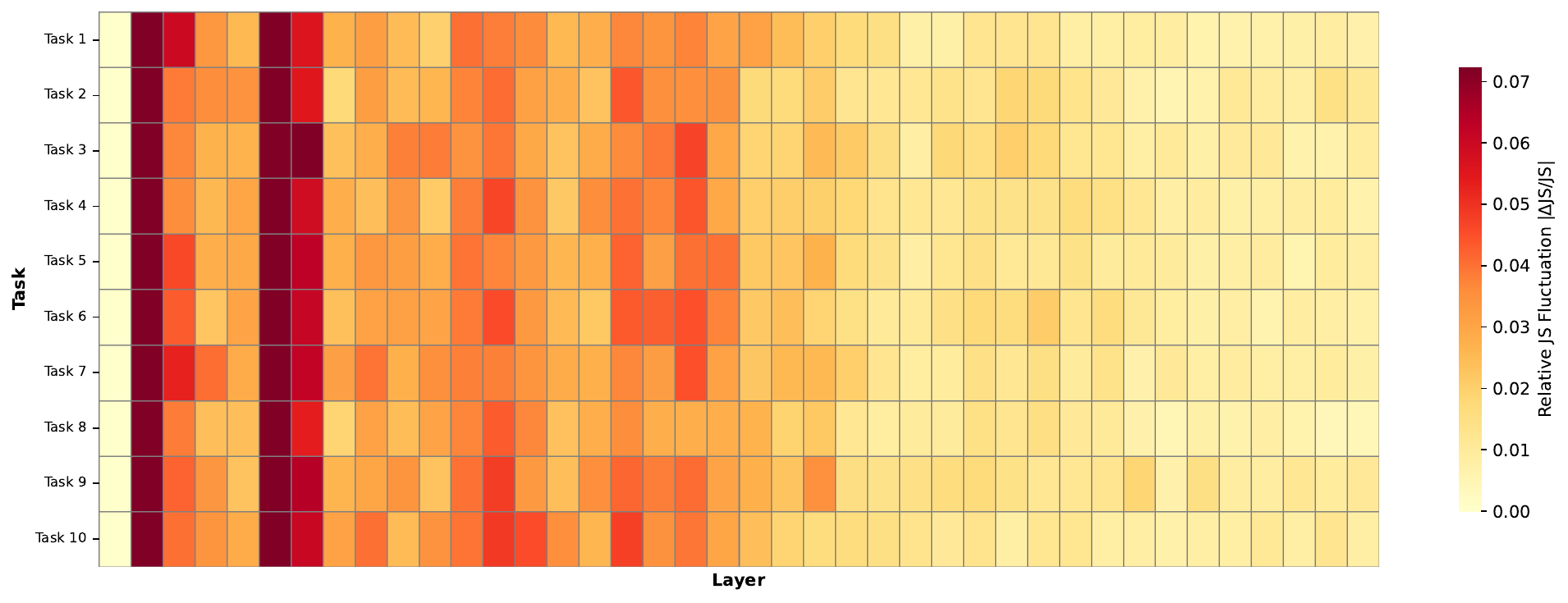}}

    \caption{
        Robustness analysis under Top-50 probability support on the AntSynNET dataset, where 500 samples are evenly divided into 10 groups of 50 samples for visualization.
        Here, Top-50 refers to the probability support used for JS-based layer-wise sensitivity computation, which is distinct from the top-10 candidate-token display in decoding.
        The single target-token probability used in $\mathrm{Ratio}$ remains unchanged.
        (a)--(c) show the variation of $\mathrm{Ratio}$ across layers for Qwen3-0.6B-Base to Qwen3-14B-Base.
        A deeper green color indicates a larger $\mathrm{Ratio}$, meaning that the target token probability changes more significantly at this layer.
        (d)--(f) show the variation of $\Delta\mathrm{JS}$ across layers under the same Top-50 setting.
        A deeper color indicates a larger abrupt change in layer sensitivity, reflecting higher vulnerability and lower robustness.
    }
    \label{fig:top50_robustness}
\end{figure*}

\section{Additional Experiments}
\label{sec:appendix_3}

\subsection{Robustness to Top-$K$ Probability Support}
\label{sec:appendix_3_1}
To examine whether the Layer-wise Sensitivity results are affected by the choice of probability support, we further repeat the analysis using a larger Top-$K$ probability support with $K=50$, as shown in Figure~\ref{fig:top50_robustness}.
For each layer, the JS divergence is computed over the aligned Top-50 probability distribution rather than a smaller truncated support.
This Top-50 setting refers to the probability support used for JS-based sensitivity computation, which is distinct from the top-10 candidate-token selection used for target-token analysis and visualization.

The results show that enlarging the probability support to Top-50 preserves the same qualitative trend observed in the main analysis: shallow layers still exhibit stronger sensitivity fluctuations, while deep layers remain comparatively stable and act as execution-dominant zones. Consequently, it further justifies our practical allocation strategy in Section~\ref{sec:Principle}, where we adopt a symmetric midpoint split to train the shallow sensitive zones and freeze the deep execution zones. 

This consistency indicates that the LayerTracer diagnosis is robust to the choice of Top-$K$ probability support, further supporting the conclusion that shallow layers are more suitable for adaptation whereas deep layers should be preserved during continued pre-training.

\subsection{Boundary Selection}
\label{sec:boundary_rationale}

\begin{figure}[t]
    \centering
    \includegraphics[width=\columnwidth]{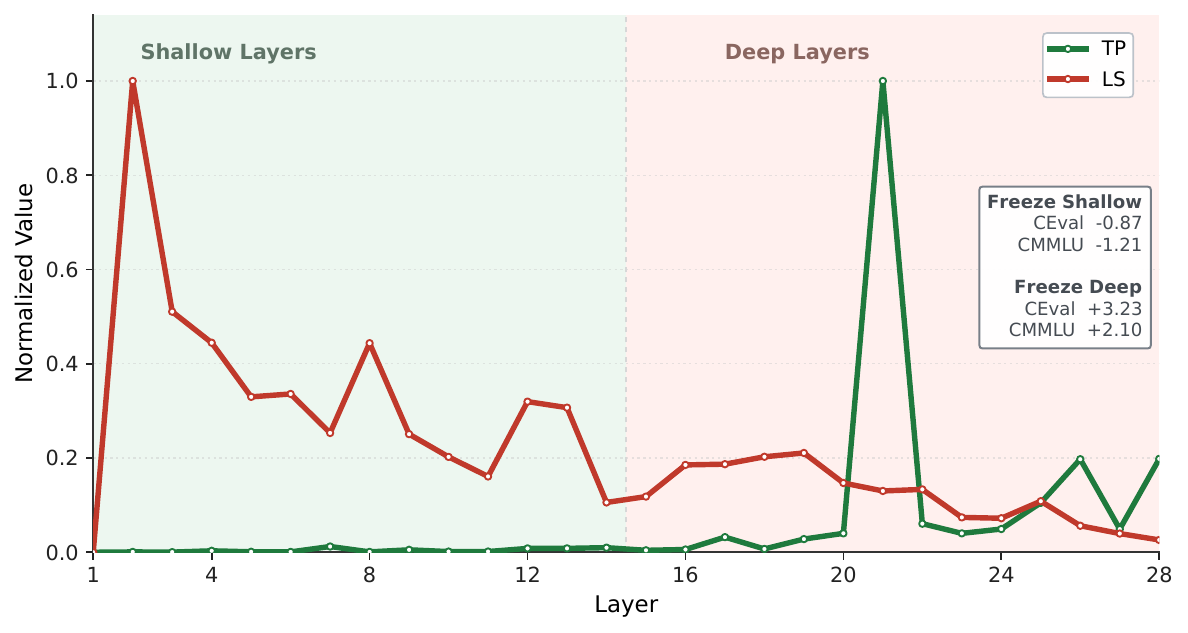}
    \caption{
    Layer-wise TP and LS profiles of Qwen3-0.6B-Base. 
    The light green region denotes the shallow layer used in the midpoint split, where LS is relatively stronger and layers are more sensitive to perturbation. 
    The light red region denotes the deep layers, where TP becomes more prominent and task evidence is mainly consolidated. 
    The dashed line marks the 50\% split used in the main experiments. 
    }
    \label{fig:evidence}
\end{figure}

An exhaustive training sweep over all possible split boundaries can identify a model-specific optimum, but this is not the goal of our study. 
Our objective is to derive a general allocation rule from layer-wise diagnostics to address \textit{RQ2}. 
Figure~\ref{fig:evidence} shows the normalized TP and LS profiles of Qwen3-0.6B-Base. 
LS is concentrated in shallow and middle layers, indicating that these layers are sensitive to perturbation and suitable for adaptation. 
In contrast, TP becomes more prominent in later layers, suggesting that deep layers mainly serve as task-evidence consolidation and execution zones.

Based on this observation, we use the 50\% midpoint as a simple and model-agnostic split in the main continued pre-training experiments. 
This setting keeps a substantial sensitive shallow layer trainable while preserving the deep execution layer from disruptive updates. 
It also provides a symmetric comparison with matched trainable-parameter budgets. 
Thus, the midpoint is not treated as an optimal boundary, but as a controlled test of the Train-Shallow/Freeze-Deep principle.

To examine whether this allocation rule depends on the midpoint, we further conduct a split-boundary diagnostic using LayerTracer. 
For a model with $N$ layers and a split ratio $r$, we set the split layer to the nearest feasible layer $b=\operatorname{round}(rN)$. 
Layers $1$ to $b$ form the shallow layer, and layers $b+1$ to $N$ form the deep layers. 
A desirable Train-Shallow/Freeze-Deep split should place high-LS layers in the trainable layer and high-TP layers in the frozen layer. 
We therefore define the following boundary alignment score:
\begin{equation}
\begin{aligned}
\mathcal{S}(b) &= \overline{\widehat{\mathrm{LS}}}_{1:b}
+ \overline{\widehat{\mathrm{TP}}}_{b+1:N} \\
&\quad -
\overline{\widehat{\mathrm{TP}}}_{1:b}
- \overline{\widehat{\mathrm{LS}}}_{b+1:N},
\end{aligned}
\end{equation}
where $\widehat{\mathrm{TP}}$ and $\widehat{\mathrm{LS}}$ denote normalized TP and LS values. 
A positive score indicates that the split is better aligned with the Train-Shallow/Freeze-Deep rule than with the reverse allocation.

As shown in Table~\ref{tab:split_boundary}, $\mathcal{S}(b)$ remains positive under 33\%, 50\%, and 66\% shallow-layer splits. 
This result indicates that the allocation direction is not tied to the exact midpoint. 
The performance deltas in Figure~\ref{fig:evidence} further support this interpretation: freezing the shallow sensitive layer and training only deep layers leads to performance drops, whereas training shallow layers while freezing deep layers yields consistent gains over full-parameter training.

\begin{table}[!t]
\centering
\small
\setlength{\tabcolsep}{5pt}
\begin{tabular}{c c c}
\toprule
\textbf{Trainable Shallow Ratio} & \textbf{Split Layer} & $\mathcal{S}(b)$ \\
\midrule
33\% & 9  & 0.004 \\
50\% & 14 & 0.138 \\
66\% & 19 & 0.392 \\
\bottomrule
\end{tabular}
\caption{
Split-boundary diagnostic based on normalized TP and LS. 
Positive scores indicate that sensitive layers are assigned to the trainable layer and task-execution layers are assigned to the frozen layer, supporting the Train-Shallow/Freeze-Deep rule across multiple split boundaries.
}
\label{tab:split_boundary}
\end{table}

\begin{table}[t]
\centering
\small
\setlength{\tabcolsep}{6pt}
\begin{tabular}{lcc}
\toprule
\textbf{Benchmark} & \textbf{Generate Std.} & \textbf{Logit Std.} \\
\midrule
C-Eval & 0.00 & 0.00 \\
CMMLU  & 0.00 & 0.00 \\
\bottomrule
\end{tabular}
\caption{
Standard deviation over five repeated evaluations for all models used in Section~\ref{sec:effective} and Section~\ref{sec:case_study}. 
}
\label{tab:eval_std}
\end{table}

\begin{table*}[t]
\centering
\small
\setlength{\tabcolsep}{4pt}
\renewcommand{\arraystretch}{1.15}
\begin{tabular}{p{0.06\linewidth} p{0.38\linewidth} p{0.50\linewidth}}
\toprule
\textbf{ID} & \textbf{Prompt} & \textbf{Response} \\
\midrule
1 &
According to traffic regulations, a motor vehicle running a red light will &
In 2019, the National Bureau of Statistics of China released the \textit{China Statistical Yearbook 2018}. The book states that \ldots ranked second, but per capita GDP was only around USD 35,000. It also notes that China's total economic output had reached first place in the world \ldots \\
\midrule
2 &
Beijing is the capital of China, while Shanghai is &
In 2019, many major changes occurred in China and around the world. These changes include: 1. China and the world \ldots the economies of China and the world are undergoing major changes. These changes include: 1. the economies of China and the world are undergoing major \ldots \\
\midrule
3 &
Washington, D.C. is the capital of the United States, while New York is &
The capital of the United States. In 1940, New York City was the capital of the United States, and also the capital of the United States. New York City \ldots capital. New York City is the capital of the United States, and also the capital of the United States \ldots \\
\bottomrule
\end{tabular}
\caption{Bad cases from Qwen3-0.6B-Base (Frozen/Trainable). The responses exhibit severe semantic drift: instead of completing simple factual prompts, the model generates unrelated news-style or encyclopedic text.}
\label{tab:case_study_outputs}
\end{table*}

\begin{table*}[t]
\centering
\small
\setlength{\tabcolsep}{4pt}
\renewcommand{\arraystretch}{1.15}
\begin{tabular}{p{0.06\linewidth} p{0.36\linewidth} p{0.52\linewidth}}
\toprule
\textbf{ID} & \textbf{Prompt} & \textbf{Response} \\
\midrule
1 &
According to traffic regulations, a motor vehicle running a red light will &
Be fined 20 yuan. If the driver does not stop in time when running the red light, an additional fine of no more than 15 yuan will be imposed.  The car I was driving hit \ldots \\
\midrule
2 &
Beijing is the capital of China, while Shanghai is &
The capital of China after 1949. Therefore, from a historical perspective, there have been two capitals in Chinese history. The first was during the Qin period, when Xianyang was established as the capital in order to strengthen centralization. \ldots \\
\midrule
3 &
Washington, D.C. is the capital of the United States, while New York is &
The capital of the United States. In the 19th century, New York City's population had already exceeded two million, making it one of the most populous cities in the world at that time. \ldots \\
\bottomrule
\end{tabular}
\caption{Good cases from Qwen3-0.6B-Base (Trainable/Frozen). Compared with the Frozen/Trainable model, the responses are more logically coherent and more likely to start with the correct completion, although some continuations still show mild topic drift.}
\label{tab:case_study_partially_corrected_outputs}
\end{table*}

\subsection{Standard Deviation under Deterministic Evaluation}
\label{sec:zero_std}

As shown in Table~\ref{tab:eval_std}, all models used in Section~\ref{sec:effective} and Section~\ref{sec:case_study} achieve zero standard deviation over five repeated evaluations on both C-Eval and CMMLU under the two evaluation protocols.
For the generation-based evaluation, we use greedy decoding with \texttt{do\_sample=False}, so the generated answer is deterministic given the same model and input. 
For the logit-based evaluation, the score is computed directly from the fixed probabilities of the candidate answer tokens, without any decoding randomness. 
Therefore, the repeated evaluations produce identical results. 
This confirms the reproducibility of our reported scores under the fixed evaluation setting.

\section{Case Study}
\label{sec:appendix_4}
To intuitively compare different layer-wise training allocation strategies, we present representative generation examples from two Qwen3-0.6B-Base variants: Qwen3-0.6B-Base (Frozen/Trainable), where shallow layers are frozen and deep layers are trainable, and Qwen3-0.6B-Base (Trainable/Frozen), where shallow layers are trainable and deep layers are frozen. 

As shown in Table~\ref{tab:case_study_outputs}, the Frozen/Trainable model often fails to follow simple factual prompts and quickly drifts into unrelated news-style or encyclopedic text. In contrast, Table~\ref{tab:case_study_partially_corrected_outputs} shows that the Trainable/Frozen model produces more coherent continuations and is more likely to begin with the correct answer, although some outputs still contain later topic drift. These qualitative cases further support our conclusion that updating deep execution layers can disrupt inherited knowledge, while freezing deep layers better preserves factual consistency and generation stability.

\end{document}